\documentclass[pdflatex,sn-mathphys-num]{sn-jnl}

\usepackage{graphicx}
\usepackage{amsmath,amssymb,amsthm}
\usepackage{bm}
\usepackage{mathrsfs}
\usepackage{xspace}
\usepackage{multirow}
\usepackage{booktabs}
\usepackage{matlab-prettifier}
\usepackage{algorithm2e}
\RestyleAlgo{ruled}
\usepackage{float}
\usepackage{placeins}
\usepackage{xcolor}
\usepackage{lineno}

\SetKwFunction{Fn}{Function}
\SetKwProg{PPR}{Function Piecewise\_regression}{}{end}%
\SetKwProg{UBP}{Function Update\_BP}{}{end}%
\SetKwProg{GPL}{Function Find\_BP}{}{end}%
\SetKwProg{OBP}{Function Refined\_BP}{}{end}%
\SetKwProg{FPPR}{Piecewise\_regression\!}{}{}%
\SetKwProg{FUBP}{Update\_BP\!}{}{}%
\SetKwProg{FGPL}{Find\_BP\!}{}{}%
\SetKwProg{FOBP}{Refined\_BP\!}{}{}%
\SetKwInOut{Input}{Input}
\SetKwInOut{Output}{Output}
\SetKwRepeat{Do}{do}{while}

\newtheorem{theorem}{Theorem}[section]

\newtheorem{remark}[theorem]{Remark}

\newtheorem{assumption}[theorem]{Assumption}

\newcommand*{\horzbar}{\rule[.5ex]{2.5ex}{0.5pt}}
\newcommand{\MATLAB}{\textsc{Matlab}\xspace}

\definecolor{red}{rgb}{1,0,0}
\definecolor{blue}{rgb}{0,0,1}

\begin{document}

\title[Improved Breakpoint Identification]{Improved identification of breakpoints in continuous piecewise regression}

\author[1]{\fnm{Taehyeong} \sur{Kim}}
\author[2]{\fnm{Hyungu} \sur{Lee}}
\author[3]{\fnm{Myungjin} \sur{Kim}}
\author*[1,2]{\fnm{Hayoung} \sur{Choi}}\email{hayoung.choi@knu.ac.kr}

\affil[1]{\orgdiv{Nonlinear Dynamics and Mathematical Application Center}, \orgname{Kyungpook National University}, \orgaddress{\city{Daegu}, \postcode{41566}, \country{Republic of Korea}}}
\affil[2]{\orgdiv{Department of Mathematics}, \orgname{Kyungpook National University}, \orgaddress{\city{Daegu}, \postcode{41566}, \country{Republic of Korea}}}
\affil[3]{\orgdiv{Department of Statistics, KNU G-LAMP Project Group, KNU Institute of Basic Sciences}, \orgname{Kyungpook National University}, \orgaddress{\city{Daegu}, \postcode{41566}, \country{Republic of Korea}}}

\abstract{
Identifying breakpoints in piecewise regression is critical for reliable and interpretable data fitting.
We propose greedy algorithms for continuous piecewise polynomial regression that search breakpoint locations over a finite, data-adaptive candidate set.
In each outer iteration, every interior breakpoint is updated by comparing three neighboring candidates (move left, stay, move right) via small local constrained least-squares (KKT) problems, which avoids step-size tuning.
With an explicit stopping rule based on fixed-point/cycle detection on the finite candidate set, the breakpoint-location search terminates in finitely many iterations.
We further combine this location update with a backward elimination scheme to select a data-driven number of breakpoints controlled by a relative MSE tolerance parameter $\tau$ and an upper bound $p$ on the number of interior breakpoints.
Experiments on synthetic and real-world datasets demonstrate competitive accuracy and computational efficiency, and the estimated breakpoints provide interpretable information about structural changes in the data.
}

\maketitle

\allowdisplaybreaks

\section{Introduction}

Piecewise regression, also known as segmented regression, is a powerful statistical technique used to model and analyze variables whose relationship can vary over different intervals of an independent variable~\citep{bai1997estimation,muggeo2003estimating}. Unlike traditional regression models, which assume a uniform effect across the entire range of data, piecewise regression allows for varying slopes and intercepts within predefined segments of the data set. This method is particularly useful in fields such as economics, epidemiology, and environmental science, where abrupt changes are often expected as a result of interventions or natural thresholds~\citep{killick2012optimal}.

Piecewise regression allows for more accurate modeling and prediction by dividing the data into segments and fitting a polynomial regression model to each segment.
There are various fields such as economics and finance~\citep{tomal2021bayesian}, engineering~\citep{gunnerud2010oil,wu2022synchronous}, energy and power systems~\citep{ding2019data,guan2018polynomial,hwangbo2018spline}, network flow~\citep{muriel2004capacitated}, biological and healthcare~\citep{greene2015improved,vieth1989fitting,wagner2002segmented} using piecewise regression. 

Continuity at breakpoints in piecewise regression is a critical factor in enhancing the reliability and interpretability of the model. 
In situations with multiple change points, continuous models can stably reflect changes in the data, thereby improving the stability of the model and predictive performance \citep{fearnhead2007line}. 
Furthermore, maintaining continuity in regression models with unknown breakpoints provides more realistic and interpretable results \citep{muggeo2003estimating}. 
Ensuring continuity in linear models with multiple structural changes more accurately captures the structural changes in the data \citep{bai1998estimating}.

The key challenge in piecewise regression is to accurately identify the points where the relationship of variables changes, known as breakpoints or change points. 
The model's accuracy heavily depends on correctly identifying these points, as they define the boundaries between intervals where different relationships apply.
The detection of breakpoints has been addressed using various statistical and computational methods. 
Traditional approaches often involve manually setting potential breakpoints based on domain knowledge or using grid search techniques to systematically evaluate a range of breakpoints~\citep{bai2003computation}. 
More advanced techniques include algorithms such as binary segmentation and dynamic programming, which are designed to optimize the fit of piecewise models by iteratively testing potential breakpoints and evaluating their impact on the accuracy of the model~\citep{chiou2019identifying, fryzlewicz2014wild, killick2012optimal, truong2020selective}. 
Despite their effectiveness, these methods can be computationally intensive and may not scale well with large datasets or complex models. 
Various methods have been proposed, including methods based on integer linear programming~\citep{yang2016mathematical}, the $\ell_1$ trend filter~\citep{kim2009ell_1}, which is a variant of the Hodrick-Prescott (H-P) filter, recently proposed adaptive piecewise regression using optimization techniques~\citep{jeong2023trend,tibshirani2014adaptive}, and a column generation-based heuristic method~\citep{tunc2021column}.

In one of the most recent papers in this field, \citep{jeong2023trend}, adaptive piecewise linear regression (APLR) is proposed.
APLR uses gradient descent to find the locations of the segments, requiring gradients of the objective function. However, gradient-based methods often face challenges in tuning hyperparameters, such as the step size (learning rate), and are sensitive to initialization, potentially leading to poor local minima.
In contrast, our proposed method searches within a predefined discrete candidate set. This derivative-free strategy avoids the complexities of step-size tuning and ensures a monotonic decrease in the objective function at each iteration among the candidate neighbors.
In addition, it avoids the difficulties of tuning hyperparameters such as learning rate in gradient descent, so more stable convergence can be expected.
The proposed method updates the segment points in the direction that necessarily reduces the MSE among the neighboring candidates explored at each step, so there is no risk of divergence or optimization failure due to the step size problem.

In this study, we present an improved algorithm for identifying breakpoints in piecewise regression using the greedy algorithm approach. A greedy algorithm is an intuitive and simple method for solving optimization problems, characterized by making locally optimal choices at each step and then finding the global optimum. Since it incrementally constructs the solution and selects the most immediately beneficial choice at each step, it is often preferred because it is simple and easy to implement compared to more complex methods such as dynamic programming. However, it is important to note that greedy algorithms do not always guarantee a globally optimal solution and can sometimes converge to a local minimum. 

In addition, determining the optimal number and location of breakpoints in a piecewise regression is important to improve the accuracy and interpretability of the model. Too many breakpoints can lead to overfitting, and not enough breakpoints can lead to underfitting of the data. The optimal number and location of breakpoints ensure that each segment of the regression model effectively represents the underlying data, providing more reliable predictions and insights.

Our main contributions in this paper are summarized in brief as follows.

\begin{itemize}
    \item We propose a greedy breakpoint-location algorithm for continuous piecewise polynomial regression on a finite candidate set, updating each interior breakpoint through three local two-interval constrained least-squares subproblems.
    \item The resulting procedure is stable in practice because it avoids step-size tuning; under a fixed-point/cycle-detection stopping rule on the finite candidate set, it terminates in finitely many iterations.
    \item We introduce a backward elimination scheme to select a data-driven number of breakpoints, controlled by an interpretable relative MSE tolerance $\tau$ and an upper bound $p$ on the number of interior breakpoints.
\end{itemize}

This paper is structured as follows. Section 2 introduces continuous piecewise polynomial regression from an optimization point of view. Section 3 presents the proposed method and algorithms. Section 4 provides theoretical analysis. Section 5 presents numerical experiments on synthetic and real datasets. Section 6 concludes with a summary and future directions.

\subsection*{Notations:}
Let $\mathbb{R}$ (resp. $\mathbb{N}$) be the set of real (resp. natural) numbers. 
For a vector $\bm{x}$, $\|\bm{x}\|_2$ denotes the Euclidean norm. The superscript $T$ denotes the transpose operator.
The mean square error (MSE) is defined as the difference between the regressor and the observed data: 
\begin{equation*}
    \operatorname{MSE}\left(\bm{\hat{y}}, \bm{y}\right) := \frac{1}{n} \left\| \hat{\bm{y}} - \bm{y} \right\|^2_2,
\end{equation*}
where $\bm{y}$ and $\hat{\bm{y}}$ in $\mathbb{R}^n$ are the observed data and the regressors, respectively.
In the following, we denote the sorted data points as $x_1, x_2, \dots, x_n$ and the corresponding observations as $y_1, y_2, \dots, y_n$. For a given subinterval $ I_j = [\xi_{j-1}, \xi_j] $, the Vandermonde matrix $ V(I_j) $ is defined as follows  
\begin{equation*}
V(I_j) = \begin{bmatrix}
1 & z_{1,j} & z_{1,j}^2 & \cdots & z_{1,j}^d \\
1 & z_{2,j} & z_{2,j}^2 & \cdots & z_{2,j}^d \\
\vdots & \vdots & \vdots & \ddots & \vdots \\
1 & z_{m_j,j} & z_{m_j,j}^2 & \cdots & z_{m_j,j}^d 
\end{bmatrix},
\end{equation*}
where $ \{z_{i,j}\} $ are the elements of $ I_j $. We denote the block-diagonal design matrix as
\begin{equation*}
X = \operatorname{diag}(V(I_1), V(I_2), \dots, V(I_k)),
\end{equation*}
and the continuity constraints are encoded by the matrix $ C_\xi $ defined such that each row imposes $ p_j(\xi_j) = p_{j+1}(\xi_j) $. For the optimization in (4), the Lagrangian multiplier associated with these constraints is denoted by $ \lambda $.

\section{Continuous piecewise polynomial regression}

First, we rigorously formulate the problem. 
Suppose that we have data $\{(x_i, y_i)\}_{i=1}^n\subset \mathbb{R}^2$, where $x_1 \le x_2 \le \cdots \le x_n$. 
We model the observed data $y_i$ as:
\begin{equation}\label{eq:noisy_data}
    y_i = f(x_i) + \varepsilon_i,
\end{equation}
where $f$ is a continuous function whose underlying trend changes at a few, distinct breakpoints. This structure implies that the trend is consistent over long intervals, each containing a sufficient number of data points for stable regression. 
And $\varepsilon_i$ is an i.i.d. Gaussian noise term with mean $0$ and standard deviation $\sigma$. 
Specifically, as shown in equation \eqref{eq:noisy_data}, we can write $y_i \sim \mathcal{N}(f(x_i), \sigma^2)$. This leads to a probability density function
\begin{equation*}
    p\bigl(y_i; f(x_i), \sigma\bigr) = \frac{1}{\sqrt{2 \pi} \sigma}\exp\!\Bigl(-\tfrac{(y_i - f(x_i))^2}{2\sigma^2}\Bigr).
\end{equation*}
Since the noise is i.i.d., the overall likelihood of our sample $\bm{y}$ given $f$ and $\sigma$ is following:
\begin{equation*}
    \ell(\bm{y};f,\sigma) = \prod_{i=1}^{n} p\bigl(y_i; f(x_i), \sigma\bigr).
\end{equation*}
Maximizing this likelihood with respect to $f$ is equivalent to minimizing the sum of squared residuals. 
Consequently, we seek $\tilde{f}$ that solves following optimization problem for given a sample $y$, the unknown continuous function $f(x)$, and $\sigma$
\begin{equation*}
    \operatorname*{minimize}_{\tilde{f}} \sum_{i=1}^{n} \bigl(\tilde{f}(x_i) - y_i\bigr)^2,
\end{equation*}
where $\tilde{f}$ is continuous estimator of the unknown function $f$. 
To properly describe $f$, a popular family of functions can be used, such as linear and polynomial functions \citep{jeong2023trend}.
Piecewise polynomials are adopted to model data with abrupt changes in trend at specific points (breakpoints), while preserving simplicity within segments.

Next, assume that the breakpoints $\xi_1, \ldots, \xi_{k-1}$, with $\xi_1 < \cdots < \xi_{k-1}$ are given for a fixed $k\geq 2$ and that each breakpoint $\xi_{j}$ does not overlap with the data $x_{i}$.
For convenience of notation, set $\xi_0 = x_1$ and $\xi_{k} = x_n$.
And define the subinterval $I_j := [\xi_{j-1}, \xi_j]$ for $j=1,\ldots, k$. 
By assumption,  each data $x_{i}$ is contained in one single subinterval. 
The goal is to construct a continuous polynomial regression of degree $d$.  
Mathematically, the piecewise polynomial regression can be formulated as a typical least squares problem with linear equality constraints as follows:
\begin{equation}\label{eq:main1}
    \begin{aligned}
    &\operatorname*{minimize}_{\theta_{ij}} \sum\limits_{j=1}^{k} \sum\limits_{x_i\in I_j} \left(p_{j}(x_i) - y_{i}\right)^2  \\
    &\text{subject to }p_{j}(\xi_j)=p_{j+1}(\xi_j), \quad j=1,\ldots,k-1,
    \end{aligned}
\end{equation}
where $p_{j}:I_j \rightarrow \mathbb{R}$ is a $d$th degree polynomial for all $j=1,\ldots,k$, given by
$p_{j}(x)= \theta_{0j} + \theta_{1j}x + \theta_{2j}x^2+\cdots + \theta_{dj}x^d$.

To rewrite \eqref{eq:main1} in a more tractable linear algebraic form, we define a Vandermonde matrix $\bm{V}(I_j)$ for each subinterval $I_j$. 
For $z\in\mathbb{R}$ denote
\begin{equation*}
    \bm{v}(z):=
\begin{bmatrix}
    1 & z & z^2 & \cdots & z^d
\end{bmatrix}^T.
\end{equation*}
Then a Vandermonde matrix~\citep{quarteroni2010numerical} of the interval $I_j$ can be written as
\begin{equation}\label{eq:V_I}
\bm{V}(I_j) := 
\begin{bmatrix}
\horzbar \, \bm{v}(z_{1,j})^T \, \horzbar \\
\horzbar \, \bm{v}(z_{2,j})^T \, \horzbar \\
  \vdots   \\
  \vdots   \\
\horzbar \, \bm{v}(z_{m_j,j})^T \, \horzbar 
\end{bmatrix},
\end{equation}
where $z_{1,j}\leq  z_{2,j} \leq \cdots \leq z_{m_j,j}$ are all elements $x_i$ in $I_j$.
 
The optimization problem \eqref{eq:main1} can be rewritten as follows:
\begin{equation}\label{eq:main2}
    \begin{aligned}
    &\operatorname*{minimize}_{\bm{\theta}} \left\| \bm{X} \bm{\theta} - \bm{y} \right\|^2_2  \\
    &\text{subject to }\bm{C}_{\bm{\xi}} \bm{\theta} = {\bm{0}} ,
    \end{aligned}
\end{equation}
where 
\begin{align*}   
\bm{X}&=
\begin{bmatrix}
 \bm{V}(I_1) & \bm{0} & \cdots & \bm{0} \\
\bm{0} &  \bm{V}(I_2) & \cdots & \bm{0} \\
\vdots & \vdots & \ddots & \vdots \\
\bm{0} & \bm{0} & \cdots &  \bm{V}(I_k) \\
\end{bmatrix},\\
\bm{\theta}&=
\begin{bmatrix}
    \theta_{01} & \theta_{11} &\cdots & \theta_{d1} & \theta_{02} & \theta_{12} &\cdots & \theta_{dk}
\end{bmatrix}^T,\\
\bm{y}&=
\begin{bmatrix}
   y_1 & y_2 &\cdots & y_n
\end{bmatrix}^T,\\
\bm{C}_{\bm{\xi}}&=
\begin{bmatrix}
    \bm{v}(\xi_1)^T & - \bm{v}(\xi_1)^T &  &  & \bm{0}  \\
     & \bm{v}(\xi_2)^T & - \bm{v}(\xi_2)^T &  &  \\
     &  & \ddots & \ddots &  \\
    \bm{0} &  &  & \bm{v}(\xi_{k-1})^T & -\bm{v}(\xi_{k-1})^T  \\
\end{bmatrix},\\
\bm{\xi} &= \begin{bmatrix}
    \xi_1 & \xi_2 & \cdots & \xi_{k-1}
\end{bmatrix}^T.
\end{align*}
The design matrix $\bm{X}$ is formulated as a block-diagonal matrix where each block corresponds to the Vandermonde matrix associated with one subinterval $I_j$. 
By concatenating these Vandermonde matrices along the block diagonal, $\bm{X}$ simultaneously captures the polynomial terms for all subintervals.
The parameter vector $\bm{\theta}$ collects all polynomial coefficients in the subintervals $k$. 
That is, $\bm{\theta}$ is formed by stacking the $d+1$ coefficients $\theta_{0j},\ldots,\theta_{dj}$ from each polynomial $p_j(\cdot)$ defined on $I_j$. 
Hence, the individual blocks of $\bm{X}$ multiply their respective portion of $\bm{\theta}$ to produce polynomial fits within each subinterval.
And the vector $\bm{y}$ holds all the observed target values $\{y_i\}_{i=1}^n$ in a single column vector, often arranged so that the entries from each subinterval are grouped together. 
Thus, when computing the residual $\bm{X}\bm{\theta} - \bm{y}$, each block of the design matrix is paired with the subset of observations corresponding to the respective subinterval.
And the matrix $\bm{C}_{\bm{\xi}}$ encodes the continuity constraints at the breakpoints $\xi_1,\dots,\xi_{k-1}$. 
Each constraint enforces that the polynomial segments of the adjacent subintervals agree in value at their shared boundary $\xi_j$. 
In particular, each row block of $\bm{C}_{\bm{\xi}}$ contains the Vandermonde vector $\bm{v}(\xi_j)^T$ applied to the coefficients of adjoining subintervals, with opposite signs ensuring the continuity condition $p_j(\xi_j) = p_{j+1}(\xi_j)$. 
Consequently, imposing $\bm{C}_{\bm{\xi}} \bm{\theta} = \bm{0}$ guarantees that the resulting piecewise polynomial function is continuous at all specified breakpoints.

Based on the linear algebra formula in \eqref{eq:main2}, the Lagrangian multiplier can incorporate linear constraints. 
Specifically, to handle the continuity condition $\bm{C}_{\bm{\xi}} \bm{\theta} = \bm{0}$ within the goal of minimizing $\|\bm{X}\bm{\theta} - \bm{y}\|_2^2$, we can naturally construct the Lagrangian function $\mathcal{L}(\bm{\theta}, \bm{\lambda})$:
\begin{equation*}
    \mathcal{L}(\bm{\theta},\bm{\lambda}) = \bm{\theta}^T\bm{X}^T \bm{X} \bm{\theta} -2\bm{y}^T \bm{X} \bm{\theta} + \bm{y}^T \bm{y} +   \bm{\lambda}^T \bm{C}_{\bm{\xi}} \bm{\theta}.
\end{equation*}
Since the optimization problem \eqref{eq:main2} is convex to find its global minimum, it is enough to
find a point where the gradient $\nabla \mathcal{L}$ vanishes.
Setting $\nabla \mathcal{L}=0$ results in the equations
\begin{equation*}
    \begin{cases}
    \cfrac{\partial \mathcal{L}}{\partial \bm{\theta}} = 2\bm{X}^T \bm{X} \bm{\theta} -2 \bm{X}^T  \bm{y}  +  \bm{C}_{\bm{\xi}}^T \bm{\lambda}  = \bm{0},\\
    \cfrac{\partial \mathcal{L}}{\partial \bm{\lambda}} = \bm{C}_{\bm{\xi}} \bm{\theta} = \bm{0}.
    \end{cases}
\end{equation*}
Then it can be rewritten in matrix form as follows.
\begin{equation}\label{eq:matrixform1}
\begin{bmatrix}
2\bm{X}^{T}\bm{X} & \bm{C}_{\bm{\xi}}^{T}\\
\bm{C}_{\bm{\xi}} & \bm{0}
\end{bmatrix}\begin{bmatrix}
\bm{\theta}\\
\bm{\lambda}
\end{bmatrix}
 = \begin{bmatrix}
2\bm{X}^{T}\bm{y} \\ \bm{0} 
\end{bmatrix}.
\end{equation}
The block matrix in equation \eqref{eq:matrixform1} is called the Karush?uhn?ucker(KKT) matrix and if it is nonsingular, the optimal solution can be found~\citep{boyd2018introduction,dell2024polynomial,dell2022generalizations}.
Then one can solve the linear system to find the solution (see Algorithm~\ref{alg:piecewise_regression}).

\begin{algorithm}[!t]
\caption{Piecewise polynomial regression}
\label{alg:piecewise_regression}
\PPR{}{
\Input{$\{(x_i, y_i)\}_{i=1}^n\subset \mathbb{R}^2$: data; $\bm{\xi}\in\mathbb{R}^{k+1}$: breakpoints; $d \in\mathbb{N}$: degree}
\Output{$\hat{y}_1,\ldots, \hat{y}_n$: regressors of observed values $y_1,\ldots,y_n$}
Construct $\bm{X}$ from $x_1,\ldots,x_n$\\
Construct $\bm{C}_{\bm{\xi}}$ from $\xi_1,\ldots, \xi_{k-1}$\\
$\bm{A}$ $\leftarrow$ $\begin{bmatrix}2\bm{X}^{T} \bm{X} & \bm{C}_{\bm{\xi}}^{T}\\ \bm{C}_{\bm{\xi}} & \bm{0} \end{bmatrix}$\\
$\bm{b}$ $\leftarrow$ $\begin{bmatrix}
2\bm{y}^{T}\bm{X} & \bm{0}  
\end{bmatrix}^T$\\
$\bm{\theta}$ $\leftarrow$  solve \eqref{eq:matrixform1} \\
$\hat{\bm{y}}$ $\leftarrow$ $\bm{X} \bm{\theta}$\\
}
\end{algorithm}

\section{Proposed method}

If the breakpoints $\bm{\xi}=[\xi_1 ~\xi_2~ \cdots~ \xi_{k-1}]^T$ are fixed, then the piecewise polynomial regression can be obtained using Algorithm~\ref{alg:piecewise_regression}. 
However, the location of breakpoints for real data is in general unknown.  
Even, the number of breakpoints is unknown. Both the number and location of breakpoints must be chosen appropriately to get a regression that fits the data.
First, assume that the number of breakpoints is given.
To find the optimal location of breakpoints $\bm{\xi}$ and the optimal coefficients of polynomials $\bm{\theta}=
[\theta_{01} ,~ \theta_{11} ,~\cdots ,~ \theta_{d1} ,~ \theta_{02} ,~ \theta_{12} ,~\cdots ,~ \theta_{dk}]^T$,
one can solve the optimization problem. 
\begin{equation}\label{eq:main3}
    \begin{aligned}
    &\operatorname*{minimize}_{\bm{\theta},~\bm{\xi}} \sum\limits_{j=1}^{k} \sum\limits_{x_i\in I_j} \left(p_{j}(x_i) - y_{i}\right)^2  \\
    &\text{subject to }p_{j}(\xi_j)=p_{j+1}(\xi_j), \quad j=1,\ldots,k-1.
    \end{aligned}
\end{equation}
where $p_{j}:I_j \rightarrow \mathbb{R}$ is a $d$th degree polynomial for all $j=1,\ldots,k$, given by
$p_{j}(x)= \theta_{0j} + \theta_{1j}x + \theta_{2j}x^2+\cdots + \theta_{dj}x^d$.

Since this optimization problem is nonconvex~\citep{jeong2023trend}, it is not easy to solve the problem. Due to the number of too many breakpoint candidates, numerical algorithms may need an extensive computational cost. To overcome this situation, our proposed method selects each breakpoint from a certain finite set.
It is important to choose a suitable set for given data.
In this paper, consider the following set for the candidates of breakpoints:
\begin{equation*}
    \mathcal{X} = \left\{\dfrac{x_i + x_{i+1}}{2} \Big| i = 1,2,\ldots,n-1 \right\}.
\end{equation*}

The reasons for setting the candidate breakpoint as the midpoint of the data points are as follows:
First, this method is data-adaptive because it generates a candidate grid that directly reflects the distribution of the data.
Second, since the breakpoint means the point where the relationship between variables changes, it is natural to assume that it is located `between' two observations.
Third, this method provides a simple yet sufficiently dense candidate set without separate parameters, ensuring that it can search for locations close to the actual breakpoint.

For convenience in explaining the proposed method for updating the breakpoints, assume that $x_1 < x_2 < \cdots < x_n$.
Therefore, the optimization problem \eqref{eq:main3} is relaxed to the following.
\begin{equation}\label{eq:main4}
    \begin{aligned}
    &\operatorname*{minimize}_{\bm{\theta},~~\bm{\xi}\subseteq \mathcal{X}} \sum\limits_{j=1}^{k} \sum\limits_{x_i\in I_j} \left(p_{j}(x_i) - y_{i}\right)^2  \\
    &\text{subject to }p_{j}(\xi_j)=p_{j+1}(\xi_j), \quad j=1,\ldots,k-1.
    \end{aligned}
\end{equation}

\begin{figure}[!t]
    \centering
    \includegraphics[width=0.56\textwidth]{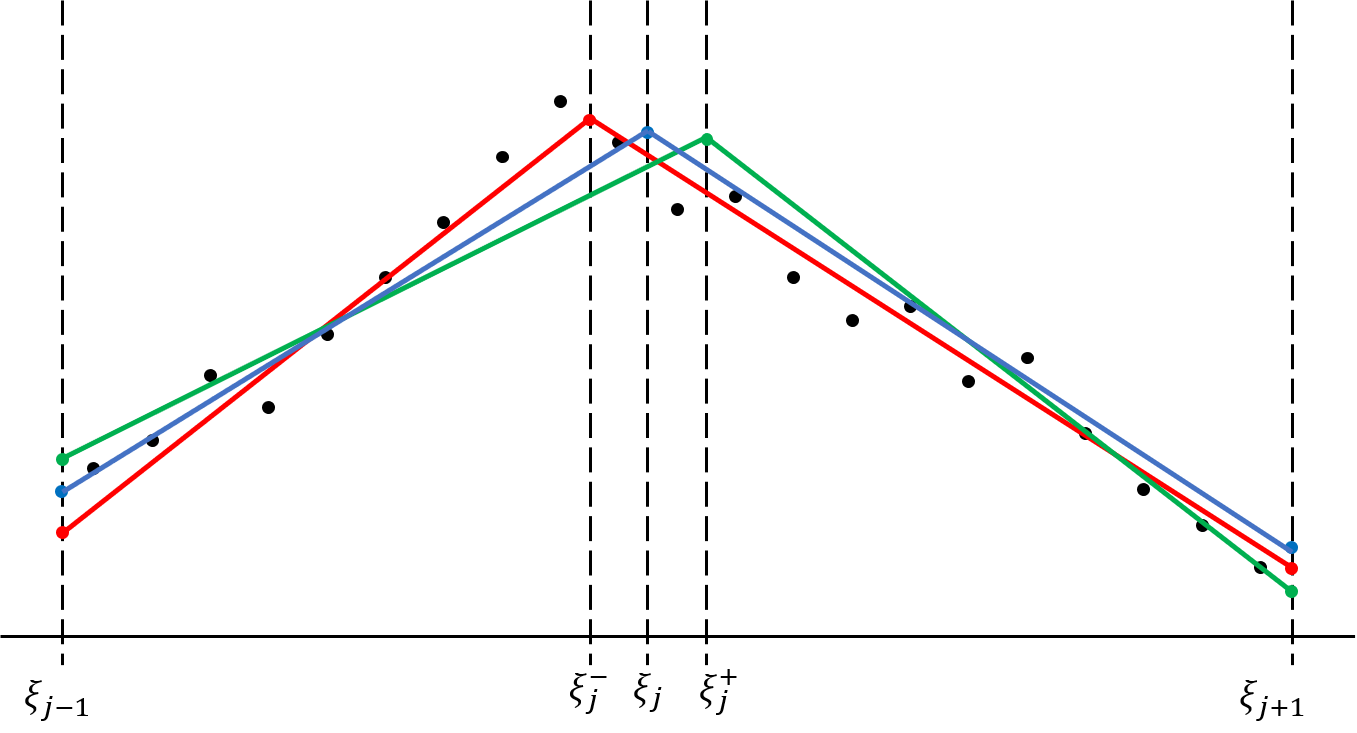}
    \caption{Update breakpoints among $\xi_{j}^{-},\xi_{j},\xi_{j}^{+}$.}
    \label{fig:proposed_method}
\end{figure}

\subsection{Update breakpoints}

For given data $\{(x_i, y_i)\}_{i=1}^n\subset \mathbb{R}^2$, where $x_1 \le x_2 \le \cdots \le x_n$ and degree of the polynomial to be regressed $d$.
To solve \eqref{eq:main1} iteratively, the sequence of breakpoint $\{\bm{\xi}^{(t)} \}$ follows the following assumptions:
\begin{assumption}\label{assump:my_assumptions}
\leavevmode
\begin{itemize}
    \item[\textbf{(A1)}] The number of segments is a predetermined integer $k$ (where $2 \le k \ll n$), which defines $k-1$ breakpoints. 
    \item[\textbf{(A2)}] For any iteration $t \ge 0$, the breakpoints $\bm{\xi}^{(t)} = (\xi_1^{(t)}, \ldots, \xi_{k-1}^{(t)})$ are strictly ordered:
    $$ x_1 = \xi_0^{(t)} < \xi_1^{(t)} < \xi_2^{(t)} < \cdots < \xi_{k-1}^{(t)} < \xi_k^{(t)} = x_n. $$
    \item[\textbf{(A3)}] The initial breakpoints $\bm{\xi}^{(0)}$ are chosen by a user-specified rule (e.g., uniform spacing or random initialization), provided that the resulting configuration satisfies \textbf{(A2)}--\textbf{(A4)}.
    \item[\textbf{(A4)}] For every iteration $t$, each interval $[\xi_{j-1}^{(t)}, \xi_j^{(t)}]$ for $j=1, \ldots, k$ must contain at least $d+1$ distinct data points.
\end{itemize}
\end{assumption}
These assumptions ensure that for all iterations $t$, the KKT matrix on the left side of \eqref{eq:matrixform1} is nonsingular.

We will explain the process of updating a single breakpoint.
For fixed $j$ and breakpoint $\xi_{j}^{(t)}$,
consider three consecutive breakpoints $\xi_{j-1}^{(t)}, \xi_{j}^{(t)}, \xi_{j+1}^{(t)}$. Then there exist two disjoint subintervals $I_j= [\xi_{j-1}^{(t)}, \xi_{j}^{(t)}]$, $I_{j+1}=[\xi_{j}^{(t)}, \xi_{j+1}^{(t)}]$. Denote all elements $x_i$ in $I_j$ and $I_{j+1}$ respectively as
\begin{align*}
    z_{1,j}&<  z_{2,j} < \cdots< z_{m_j-1,j} < z_{m_j,j},\\
   z_{1,j+1}&<  z_{2,j+1} < \cdots< z_{m_{j+1}-1,j+1} < z_{m_{j+1},j+1},
\end{align*}
where $m_j$ (resp. $m_{j+1}$) is the number of data points in $I_j$ (resp. $I_{j+1}$).

The candidate of breakpoints $\xi_{j}^{-}< \xi_{j}^{(t)}<\xi_{j}^{+}$ can be positioned as follows:
\begin{equation*}
    \begin{cases}
        \xi_{j}^{-} := \cfrac{z_{m_j-1,j} + z_{m_j,j}}{2},\\
        \xi_{j}^{(t)}: = \cfrac{z_{m_j,j} + z_{1,j+1}}{2},\\
        \xi_{j}^{+}: = \cfrac{z_{1,j+1} + z_{2,j+1}}{2},
    \end{cases}
\end{equation*}

\begin{algorithm}[!t]
\caption{Update a single breakpoint}
\label{alg:update_break_point}
\UBP{}{
\Input{$(\bm{x}_{\{j,j+1\}}, \bm{y}_{\{j,j+1\}})\in\mathbb{R}^{(m_{j}+m_{j+1})\times2}$: data;  $[\xi_{j-1}^{(t)},\xi_{j}^{(t)},\xi_{j+1}^{(t)}]\in\mathbb{R}^{3}$: breakpoints; $d \in\mathbb{N}$: degree}
\Output{$\xi_{j}^{(t+1)}\in\mathbb{R}$: updated breakpoint}
Create $\bm{V}_{j}^{-},\bm{V}_{j}^{},\bm{V}_{j}^{+}$ from $\bm{x}_{\{j,j+1\}}$\\
$f_{-}\leftarrow$ candidate $\xi_j^{-}$ satisfies Assumption~\ref{assump:my_assumptions} (\textbf{A2}) and (\textbf{A4}) locally\\
$f_{+}\leftarrow$ candidate $\xi_j^{+}$ satisfies Assumption~\ref{assump:my_assumptions} (\textbf{A2}) and (\textbf{A4}) locally\\
\uIf{$f_{-}$}{
    $\bm{\theta}^{-}_{j}$ $\leftarrow$ Solve \eqref{opti1}\\
    $r_{-}$ $\leftarrow$ $\operatorname{MSE} (\bm{X}_{j}^{-} \bm{\theta}^{-}_{j}, \bm{y}_{\{j,j+1\}})$ \\
}
\Else{
    $r_{-} \leftarrow +\infty$\\
}
$\bm{\theta}^{(t)}_{j}$ $\leftarrow$ Solve \eqref{opti2}\\
$r$ $\leftarrow$ $\operatorname{MSE} (\bm{X}_{j}^{ } \bm{\theta}^{(t)}_{j}, \bm{y}_{\{j,j+1\}})$\\
\uIf{$f_{+}$}{
    $\bm{\theta}^{+}_{j}$ $\leftarrow$ Solve \eqref{opti3}\\
    $r_{+}$ $\leftarrow$ $\operatorname{MSE} (\bm{X}_{j}^{+} \bm{\theta}^{+}_{j}, \bm{y}_{\{j,j+1\}})$\\
}
\Else{
    $r_{+} \leftarrow +\infty$\\
}
\uIf{$r_{-} < r~\mathbf{ and }~r_{-} < r_{+}$}{
    $\xi_{j}^{(t+1)}$ $\leftarrow$ $\xi_{j}^{-}$ \;
  }
  \uElseIf{$r_{+} < r_{-}~\mathbf{ and }~r_{+} < r$}{
    $\xi_{j}^{(t+1)}$ $\leftarrow$ $\xi_{j}^{+}$ \;
  }
  \Else{
    $\xi_{j}^{(t+1)}$ $\leftarrow$ $\xi_{j}^{(t)}$ \;
  }
}
\end{algorithm}

Update $\xi_{j}^{(t+1)}$ with the best among \emph{feasible} candidates in $\{\xi_{j}^{-}, \xi_{j}^{(t)}, \xi_{j}^{+}\}$ that produces the smallest MSE (see Figure~\ref{fig:proposed_method}). 
To find the best breakpoint $\xi_{j}^{(t+1)}$, it solves up to three optimization problems.
\begin{equation}\label{opti1}
    \begin{aligned}
        \mathscr{P}_{-}:~&\operatorname*{minimize}_{\bm{\theta}^{-}_{j}} \left\| \bm{X}_{j}^{-} \bm{\theta}^{-}_{j} - \bm{y}_{\{j,j+1\}} \right\|_2^2  \\
        &\text{subject to } \begin{bmatrix} \bm{v}(\xi_{j}^{-})^T & -\bm{v}(\xi_{j}^{-})^T \end{bmatrix} \bm{\theta}^{-}_{j} = 0,
    \end{aligned}
\end{equation}
\begin{equation}\label{opti2}
    \begin{aligned}
         \mathscr{P}_{0}:~&\operatorname*{minimize}_{\bm{\theta}^{(t)}_{j}} \left\| \bm{X}_{j}^{ } \bm{\theta}^{(t)}_{j} - \bm{y}_{\{j,j+1\}} \right\|_2^2  \\
        &\text{subject to } \begin{bmatrix} \bm{v}(\xi_{j}^{(t)})^T & -\bm{v}(\xi_{j}^{(t)})^T \end{bmatrix} \bm{\theta}^{}_{j} = 0,
    \end{aligned}
\end{equation}
\begin{equation}\label{opti3}
    \begin{aligned}
         \mathscr{P}_{+}:~&\operatorname*{minimize}_{\bm{\theta}^{+}_{j}} \left\| \bm{X}_{j}^{+} \bm{\theta}^{+}_{j} - \bm{y}_{\{j,j+1\}} \right\|_2^2  \\
        &\text{subject to } \begin{bmatrix} \bm{v}(\xi_{j}^{+})^T & -\bm{v}(\xi_{j}^{+})^T \end{bmatrix} \bm{\theta}^{+}_{j} = 0,
    \end{aligned}
\end{equation}
where 
\begin{align*}
\bm{X}_{j}^{-}
=&
\begin{bmatrix}
    \bm{V}\left([\xi_{j-1}^{(t)}, \xi_{j}^{-}]\right) & \\
    & \bm{V}\left([\xi_{j}^{-}, \xi_{j+1}^{(t)}]\right)
\end{bmatrix},\\
\bm{X}_{j}^{}
=&
\begin{bmatrix}
    \bm{V}\left([\xi_{j-1}^{(t)}, \xi_{j}^{(t)}]\right) & \\
    & \bm{V}\left([\xi_{j}^{(t)}, \xi_{j+1}^{(t)}]\right)
\end{bmatrix},\\
\bm{X}_{j}^{+}
=&
\begin{bmatrix}
    \bm{V}\left([\xi_{j-1}^{(t)}, \xi_{j}^{+}]\right) & \\
    & \bm{V}\left([\xi_{j}^{+}, \xi_{j+1}^{(t)}]\right)
\end{bmatrix},
\end{align*}
$\bm{V}(I)$ is the Vandermonde matrix of the interval $I$ defined in \eqref{eq:V_I} and $\bm{y}_{\{j,j+1\}}$ is the data vector of $x_i$ in $I_j \cup I_{j+1}$. 

The breakpoint $\xi_{j}^{(t)}$ can be updated as:
\begin{equation*}
    \xi_{j}^{(t+1)} = \begin{cases} 
    \xi_{j}^{-} & \text{if } r_{-} < r \text{ and } r_{-} < r_{+}, \\
    \xi_{j}^{(t)} & \text{if } r \leq r_{-} \text{ and } r \leq r_{+}, \\
    \xi_{j}^{+} & \text{if } r_{+} < r \text{ and } r_{+} < r_{-},
    \end{cases}
\end{equation*}
where
$r_{-} = \operatorname{MSE} (\bm{X}_{j}^{-} \bm{\theta}^{-}_{j}, \bm{y}_{\{j,j+1\}})$,
$r = \operatorname{MSE} (\bm{X}_{j}^{(t)} \bm{\theta}^{(t)}_{j}, \bm{y}_{\{j,j+1\}})$,
$r_{+} = \operatorname{MSE} (\bm{X}_{j}^{+} \bm{\theta}^{+}_{j}, \bm{y}_{\{j,j+1\}})$
(see Algorithm~\ref{alg:update_break_point}); for infeasible candidates, the corresponding local value is set to $+\infty$.

\begin{algorithm}[!t]
\caption{Greedy search for breakpoint locations}
\label{alg:greedy_regression}
\GPL{}{
\Input{$(\bm{x}, \bm{y})\in \mathbb{R}^{n\times 2}$: data; $\bm{\xi}^{(0)}\in \mathbb{R}^{k+1}$: initial breakpoints (including endpoints $\xi_0,\xi_k$); $d\in\mathbb{N}$: degree}
\Output{$\hat{\bm{y}}\in \mathbb{R}^{n}$: piecewise regression; $\hat{\bm{\xi}}\in \mathbb{R}^{k+1}$: best-so-far breakpoints}
$t \leftarrow 0$\\
$\hat{\bm{\xi}} \leftarrow \bm{\xi}^{(0)}$\\
$\hat{\bm{y}}$ $\leftarrow$ \FPPR{$(\bm{x}, \bm{y}, \bm{\xi}^{(t)} ,d)$}{}
$error \leftarrow \operatorname{MSE}\left(\hat{\bm{y}}, \bm{y}\right)$\\
$\mathcal{V} \leftarrow \{\bm{\xi}^{(0)}\}$\\
\While{true}{
    \For{$j\gets 1$ \KwTo $k-1$}{
    $\bm{x}_{\{j,j+1\}} \leftarrow \bm{x}(\bm{x} \geq \xi^{(t)}_{j-1} \,\,\,\&\,\,\, \bm{x} \leq \xi^{(t)}_{j+1})$\\
    $\bm{y}_{\{j,j+1\}} \leftarrow \bm{y}(\bm{x} \geq \xi^{(t)}_{j-1} \,\,\,\&\,\,\, \bm{x} \leq \xi^{(t)}_{j+1})$\\
    $\xi^{(t+1)}_{j}$ $\leftarrow$ \FUBP{$(\bm{x}_{\{j,j+1\}}, \bm{y}_{\{j,j+1\}},[\xi^{(t)}_{j-1}, \xi^{(t)}_{j}, \xi^{(t)}_{j+1}],d)$}{}
    }
    $\bm{\xi}^{(t+1)} \leftarrow [\xi_{0}, \xi^{(t+1)}_{1}, \xi^{(t+1)}_{2}, \ldots , \xi^{(t+1)}_{k-1}, \xi_{k}]$\\
    \If{$\bm{\xi}^{(t+1)} \in \mathcal{V}$}{
        \textbf{break}\;
    }
    $\mathcal{V} \leftarrow \mathcal{V} \cup \{\bm{\xi}^{(t+1)}\}$\\
    $\hat{\bm{y}}$ $\leftarrow$ \FPPR{$(\bm{x}, \bm{y}, \bm{\xi}^{(t+1)},d)$}{}
    \uIf{$\operatorname{MSE}\left(\hat{\bm{y}}, \bm{y}\right) < error$}{
        $error \leftarrow \operatorname{MSE}\left(\hat{\bm{y}}, \bm{y}\right)$\\
        $\hat{\bm{\xi}} \leftarrow \bm{\xi}^{(t+1)}$ \;
    }
    $t \leftarrow t+1$\\ 
}
$\hat{\bm{y}}$ $\leftarrow$ \FPPR{$(\bm{x}, \bm{y}, \hat{\bm{\xi}},d)$}{}
}
\end{algorithm}

\medskip

    
    

\begin{algorithm}[!t]
\caption{Refining the number of breakpoints (backward elimination)}
\label{alg:optimal_BP_algorithm}
\OBP{}{
\Input{$(\bm{x}, \bm{y})\in\mathbb{R}^{n\times 2}$: data; $\bm{\xi} \in\mathbb{R}^{k+1}$: initial breakpoints (including endpoints $\xi_0,\xi_k$); $d\in \mathbb{N}$: degree; \\$\tau \geq 1$: relative MSE tolerance for removing one breakpoint;\\ $p\in\mathbb{N}$: upper bound on the number of interior breakpoints}
\Output{$\check{\bm{\xi}}$: selected breakpoints}
\While{true}{
    $[\hat{\bm{y}}, \hat{\bm{\xi}}]$ $\leftarrow$ \FGPL{$(\bm{x}, \bm{y},\bm{\xi},d)$}{}
    $base \leftarrow \operatorname{MSE}\left(\hat{\bm{y}}, \bm{y}\right)$\\
    $\rho_{\min} \leftarrow \infty$\\
    \For{$j\gets 1$ \KwTo $\operatorname{length}(\hat{\bm{\xi}})-2$}{
        $\bm{\zeta}_j \leftarrow$ Remove the $j$-th interior breakpoint from $\hat{\bm{\xi}}$\\
        $\tilde{\bm{y}}$ $\leftarrow$ \FPPR{$(\bm{x}, \bm{y}, \bm{\zeta}_j ,d)$}{}
        $\rho_j \leftarrow \cfrac{\operatorname{MSE}\left(\tilde{\bm{y}}, \bm{y}\right)}{base}$\\
        \If{$\rho_j \leq \rho_{\min}$}{
            $\rho_{\min} \leftarrow \rho_j$\\
            $\bm{\xi}_{\mathrm{next}} \leftarrow \bm{\zeta}_j$
        }
    }
    \If{$\rho_{\min} \geq \tau$ \textbf{or} $\operatorname{length}(\hat{\bm{\xi}})-2 \le p$}{
        $\check{\bm{\xi}} \leftarrow \hat{\bm{\xi}}$\\
        \textbf{break}\;
    }
    $\bm{\xi} \leftarrow \bm{\xi}_{\mathrm{next}}$\\
}
}
\end{algorithm}

Note that Algorithm~\ref{alg:update_break_point} updates only a single breakpoint. 
To update all breakpoints, the algorithm repeats this process consecutively for $j=1,2,\ldots,k-1$. 
The first $\xi_{0}$ and the last $\xi_{k}$ are not updated for any iteration.
For each iteration, the breakpoints $\xi_{1}^{(t)}, \ldots , \xi_{k-1}^{(t)}$ are updated independently.
This means that the updated $\xi_{j}^{(t+1)}$ has no effect on the updating process of $\xi_{1}^{(t)},\ldots,\xi_{j-1}^{(t)},\xi_{j+1}^{(t)}\ldots,\xi_{k-1}^{(t)}$.
The initial breakpoints are chosen randomly or heuristically. 
There are two criteria under which Algorithm~\ref{alg:greedy_regression} terminates. 
The first is when all breakpoints are no longer updated, and the second is when an updated breakpoint appeared in the previous iteration (see  Algorithm~\ref{alg:greedy_regression}).

\medskip

The main computational complexity of Algorithm~\ref{alg:update_break_point} lies in solving
optimization problems \eqref{opti1} through \eqref{opti3}.
Each optimization problem within a subinterval can be addressed using Algorithm~\ref{alg:update_break_point}, which involves computing the inverse of a $(2d+3)\times(2d+3)$ matrix as described in \eqref{eq:matrixform1}, where $d$ is the polynomial degree.
In particular, for piecewise linear regression, each optimization problem becomes the problem of finding the inverse of a matrix of size $5\times 5$, which is a very time-efficient problem. 
Moreover, parallel computation technique can be applied since the updated breakpoint $\xi^{(t+1)}_{j}$ for the $j$-th breakpoint $\xi^{(t)}_{j}$ in the $t$-th iteration is not reflected in the update of the ($j+1$)-th breakpoint $\xi^{(t)}_{j+1}$ in the $t$-th iteration.

\subsection{Finding the optimal number of breakpoints}

In this section, we propose an algorithm to determine a suitable set of breakpoints for piecewise polynomial regression using a backward elimination strategy.
Selecting an appropriate number of breakpoints is crucial for accurately modeling and interpreting the data, balancing model complexity and goodness-of-fit. 
The number of breakpoints is closely related to the under-fitting and over-fitting of piecewise regression models. Too few breakpoints can lead to a model that misses important trend changes, while too many can result in a model that is overly complex and sensitive to random fluctuations in the data~\citep{jeong2023trend}. 
By finding the optimal number of breakpoints, a balance can be achieved that provides meaningful insights while maintaining model parsimony.

\begin{figure*}[!ht]
    \centering
    \includegraphics[width=0.8\textwidth]{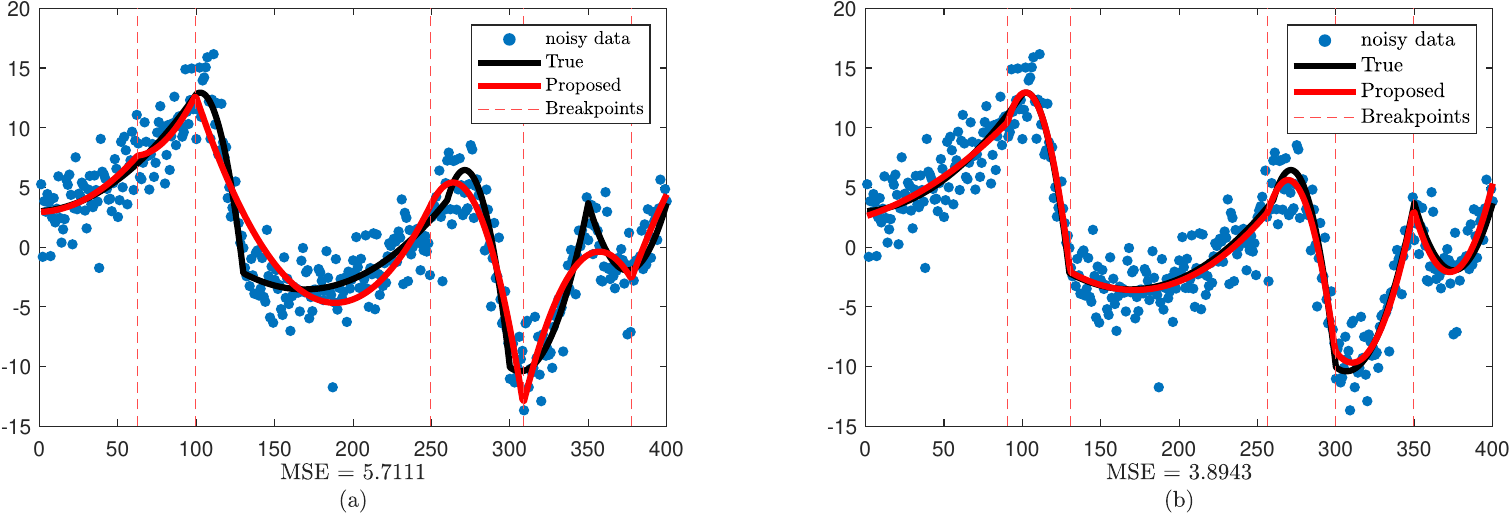}
    \caption{Algorithm~\ref{alg:optimal_BP_algorithm} can avoid local minimum well.}
    \label{fig:quadratic}
\end{figure*}

Similarly to other iteration-based algorithms for nonconvex optimization, Algorithm~\ref{alg:greedy_regression} may possibly converge to a local minimum depending on the initial value (see Figure~\ref{fig:quadratic} (a)). 
To avoid this situation, Algorithm~\ref{alg:optimal_BP_algorithm} starts 
with a sufficiently large number of breakpoints and then removes the most redundant breakpoint in the sense of optimization. 

Specifically, assume that breakpoints $\bm{\xi}= [\xi_0 ~\xi_1 ~ \cdots ~ \xi_k]^T$ with endpoints $\xi_0$, $\xi_k$ are found by Algorithm~\ref{alg:greedy_regression}.
Consider the following $k-1$ optimization problems for $i = 1,2,\ldots,k-1$.
This optimization problem is equivalent to piecewise polynomial regression in the absence of the $i$-th breakpoint in $\bm{\xi}$.
And then select the most redundant breakpoint $\xi_{i}$, which has the smallest MSE among the optimization problems $\mathscr{P}_{i}$.

\begin{equation}\label{opti_merge}
    \begin{aligned}
         \mathscr{P}_{i}:~ &\operatorname*{minimize}_{\bm{\theta}^{}_{i}} \left\| \bm{X}_{i}^{} \bm{\theta}^{}_{i} - \bm{y}_{} \right\|_2^2  \\
        &\text{subject to } \bm{C}_{\bm{\zeta}_i}\bm{\theta}^{}_{i} = \bm{0}
    \end{aligned}
\end{equation}
where 
\begin{align*}   
\bm{X}_{i}&=
\begin{bmatrix}
    \bm{V}(I_1) &  & \cdots &  &  \bm{0}\\
     & \ddots &  &  &  \\
    \vdots &  & \bm{V}(I_{i-1} \cup I_{i} ) &  &  \vdots\\
     &  &  & \ddots &  \\
    \bm{0} &  & \cdots &  &  \bm{V}(I_k)
\end{bmatrix},\\
\bm{C}_{\bm{\zeta}_i}&=
\begin{bmatrix}
\bm{v}(\xi_1)^T & - \bm{v}(\xi_1)^T \\
& \ddots & \ddots &    \\
& & \bm{v}(\xi_{i-1})^T & -\bm{v}(\xi_{i-1})^T \\
& &  & \bm{v}(\xi_{i+1})^T & -\bm{v}(\xi_{i+1})^T \\
& & &  & \ddots & \ddots  \\
& & & &  & \bm{v}(\xi_{k-1})^T  &  -\bm{v}(\xi_{k-1})^T
\end{bmatrix},\\
\bm{\zeta}_i &= 
   \begin{bmatrix}\xi_{0} & \xi_{1} & \cdots & \xi_{i-1} & \xi_{i+1} & \cdots & \xi_{k-1} & \xi_{k} \end{bmatrix}^T.
\end{align*}

To determine which breakpoints to remove, let $\bm{\theta}$ be a solution of the optimization problem \eqref{eq:main2} with breakpoint $\bm{\xi}$. Consider the following set: 
\begin{equation*}
    \mathcal{S} = \left\{
    \frac{\operatorname{MSE}(\bm{X}_i\bm{\theta}_i,\bm{y})}{\operatorname{MSE}(\bm{X}\bm{\theta},\bm{y})} \Big| \bm{\theta}_i : \text{optimal solution of }\mathscr{P}_i
    \right\}.
\end{equation*}
Then the elements of $\mathcal{S}$ represent the ratio of MSE for the piecewise regression models before and after removal of the breakpoint $\xi_i$. 
The breakpoint $\xi_i$ which minimizes $\mathcal{S}$ represents that it has the least impact on the results of the piecewise regression.
If $\min{\mathcal{S}}$ is sufficiently large, this indicates that removing any breakpoints significantly affects the piecewise regression, suggesting that all breakpoints are critical to the model.
For each candidate removal $\bm{\zeta}_i$ obtained by deleting the $i$-th interior breakpoint from $\hat{\bm{\xi}}$, we evaluate the MSE ratio using \texttt{Piecewise\_regression} with the remaining breakpoints fixed. After selecting the breakpoint to remove, we run Algorithm~\ref{alg:greedy_regression} again on the reduced breakpoint set to further refine the remaining breakpoint locations.
This process is repeated until the stopping criterion is met (see Algorithm~\ref{alg:optimal_BP_algorithm}).

\begin{figure*}[!ht]
    \centering
    \includegraphics[width=0.8\textwidth]{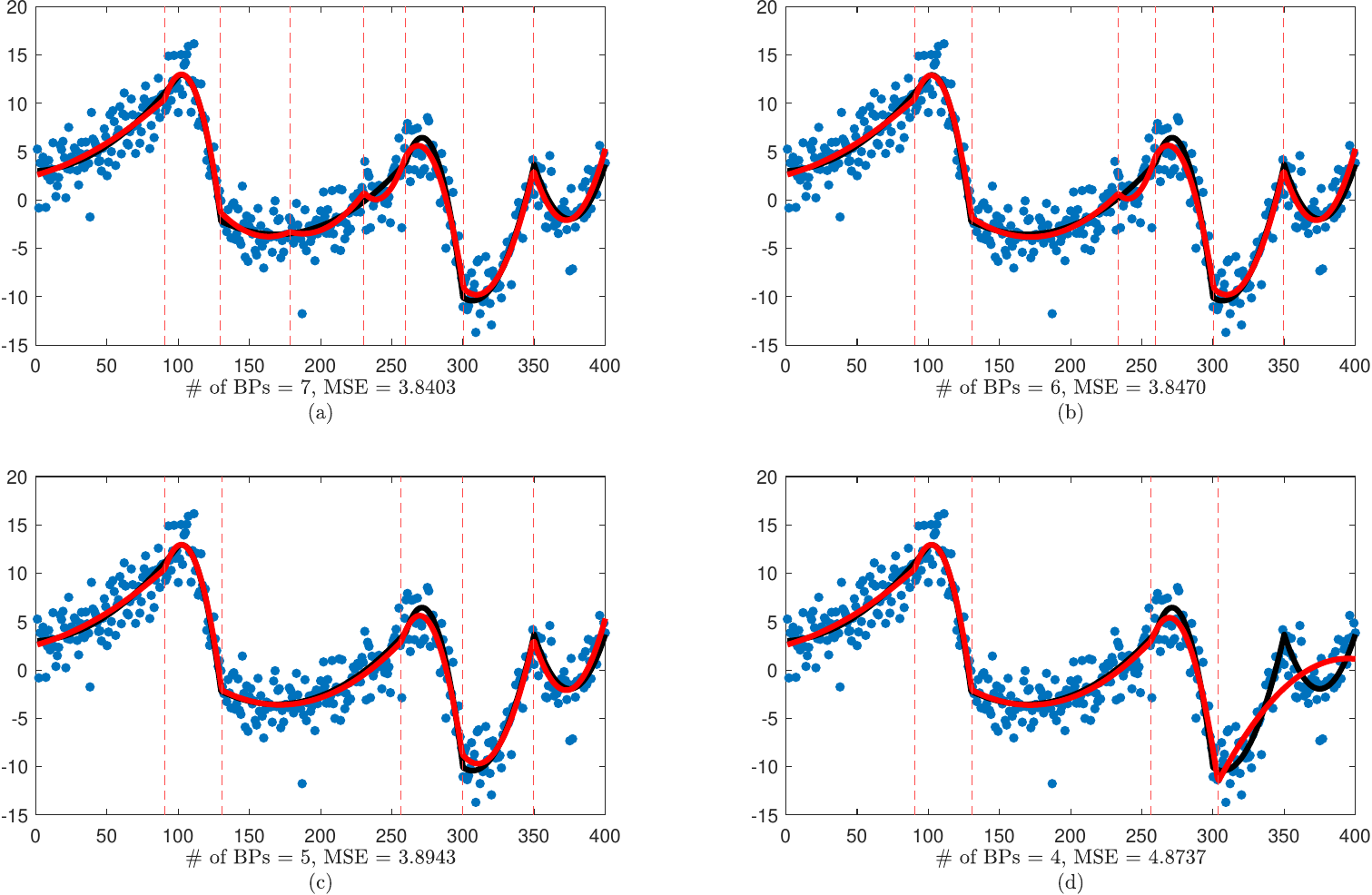}
    \caption{Comparing MSE as the number of breakpoints is reduced from $7$ to $4$.} 
    \label{fig:BPs}
\end{figure*}

There are two stopping criteria for Algorithm~\ref{alg:optimal_BP_algorithm}.
The first stopping criterion is the tolerance $\tau \geq 1$, which controls the allowed \emph{relative} increase in MSE when a breakpoint is removed (see the ratio set $\mathcal{S}$).
If $\min \mathcal{S} \ge \tau$, then removing any single interior breakpoint increases the MSE by at least a factor $\tau$, and the current breakpoint set is accepted and the elimination stops.
The tolerance $\tau$ controls the balance between overfitting and underfitting in piecewise regression.
A higher value of $\tau$ implies a wider tolerance range, leading to the removal of more breakpoints, which can result in underfitting.
On the other hand, a lower $\tau$ indicates a narrower tolerance range, leading to the retention of more breakpoints, thereby increasing the risk of overfitting.
Therefore, selecting an appropriate value of $\tau$ is crucial and should be based on the characteristics of the data.

The second criterion is $p \in \mathbb{N}$, representing an upper bound on the number of \emph{interior} breakpoints (excluding the two endpoints $\xi_0,\xi_k$).
The process of removing breakpoints is repeated until the number of interior breakpoints is equal to or less than $p$.
This criterion is particularly effective when prior knowledge about the data suggests a specific breakpoint budget.

For example, the MSE of the piecewise regression using Algorithm~\ref{alg:optimal_BP_algorithm}, starting with $7$ breakpoints and reducing the breakpoints by one, is shown in Figure~\ref{fig:BPs}. The MSE increases very slightly when there are enough breakpoints by eliminating inconsequential breakpoints, but it increases dramatically from $3.8943$ to $4.8737$ when the number of breakpoints was reduced from $5$ to $4$. With the tolerance rule controlled by $\tau$, the algorithm stops and selects $5$ breakpoints because further removal would cause a noticeable increase in MSE.

Furthermore, Figure~\ref{fig:quadratic} shows the effectiveness of the proposed Algorithm~\ref{alg:optimal_BP_algorithm}.
Figure \ref{fig:quadratic}.(a) is the result of Algorithm~\ref{alg:greedy_regression} for five randomly generated initial breakpoints $\xi_1 = 69.5, \xi_2 = 99.5, \xi_3 = 239.5, \xi_4 = 319.5, \xi_5 = 369.5$.
The MSE is $5.7111$, indicating that the piecewise regression reaches a local minimum and does not fit the data well. 
However, the MSE of the regression in Figure \ref{fig:quadratic}~(b), which is the result of Algorithm~\ref{alg:optimal_BP_algorithm}, is $3.8943$, effectively avoiding the local minimum.
This even has a lower MSE than the true piecewise quadratic model before adding noise, which has an MSE of $4.0988$, represented by the solid black line.

\section{Theoretical Analysis}\label{sec:theory}

In this section, we establish the theoretical foundations of the proposed algorithm.
We first prove that the constrained least-squares subproblem \eqref{eq:main2} is well posed for any admissible breakpoint configuration.
We then show that Algorithm~\ref{alg:greedy_regression} terminates in finite time under an explicit stopping rule (fixed point or cycle detection) on the finite candidate set.
We also clarify what kind of ``local optimality'' can be concluded from Algorithm~\ref{alg:update_break_point}: it is a coordinate-wise stationarity statement with respect to the \emph{local two-interval surrogate problems} \eqref{opti1}--\eqref{opti3}, not necessarily a coordinate-wise local optimum for the \emph{global} MSE objective unless additional assumptions are imposed.
Throughout, we assume that $x_1 < x_2 < \cdots < x_n$ and that every breakpoint configuration satisfies Assumption~\ref{assump:my_assumptions}.

The following theorem guarantees the existence and uniqueness of the solution to the constrained least-squares problem for any fixed admissible breakpoint vector.

\begin{theorem}\label{thm:well_posed}
    Let $\bm{\xi} = (\xi_1, \ldots, \xi_{k-1})$ be an admissible breakpoint vector satisfying Assumption~\ref{assump:my_assumptions}.
    Then the KKT matrix
    \begin{equation*}
        \bm{K}(\bm{\xi}) = \begin{bmatrix} 2\bm{X}(\bm{\xi})^T \bm{X}(\bm{\xi}) & \bm{C}_{\bm{\xi}}^T \\ \bm{C}_{\bm{\xi}} & \bm{0} \end{bmatrix}
    \end{equation*}
    in \eqref{eq:matrixform1} is nonsingular, and consequently the constrained least-squares problem \eqref{eq:main2} has a unique solution.
\end{theorem}

\begin{proof}
    We show that $\bm{K}(\bm{\xi})$ has trivial nullspace.
    
    Consider a segment $I_j = [\xi_{j-1}, \xi_j]$ containing $m_j$ data points $z_{1,j} < z_{2,j} < \cdots < z_{m_j,j}$.
    By Assumption~\ref{assump:my_assumptions}(\textbf{A4}), we have $m_j \ge d+1$.
    Suppose $\bm{V}(I_j) \bm{a} = \bm{0}$ for some $\bm{a} = (a_0, \ldots, a_d)^T \in \mathbb{R}^{d+1}$.
    Then the polynomial $q(z) := \sum_{\ell=0}^{d} a_\ell z^\ell$ satisfies $q(z_{i,j}) = 0$ for $i = 1, \ldots, m_j$.
    Since $q$ has degree at most $d$ but vanishes at $m_j \ge d+1$ distinct points, we must have $q \equiv 0$, hence $\bm{a} = \bm{0}$.
    This proves that each Vandermonde block $\bm{V}(I_j)$ has full column rank.
    
    Since the design matrix $\bm{X}(\bm{\xi})$ is block-diagonal with blocks $\bm{V}(I_1), \ldots, \bm{V}(I_k)$, it follows that $\bm{X}(\bm{\xi})$ has full column rank, and hence $\bm{H}(\bm{\xi}) := 2\bm{X}(\bm{\xi})^T \bm{X}(\bm{\xi})$ is symmetric positive definite.
    
    Next, we show that the constraint matrix $\bm{C}_{\bm{\xi}} \in \mathbb{R}^{(k-1) \times k(d+1)}$ has full row rank.
    Let $\bm{r}_j^T$ denote the $j$-th row of $\bm{C}_{\bm{\xi}}$, which has the form $[\bm{0}, \ldots, \bm{v}(\xi_j)^T, -\bm{v}(\xi_j)^T, \ldots, \bm{0}]$ with nonzero blocks only in positions $j$ and $j+1$.
    Suppose $\sum_{j=1}^{k-1} \alpha_j \bm{r}_j^T = \bm{0}^T$ for some scalars $\alpha_1, \ldots, \alpha_{k-1}$.
    Examining the first block gives $\alpha_1 \bm{v}(\xi_1)^T = \bm{0}^T$, and since $\bm{v}(\xi_1) \neq \bm{0}$, we have $\alpha_1 = 0$.
    Proceeding inductively, if $\alpha_1 = \cdots = \alpha_{j-1} = 0$, then examining the $j$-th block gives $\alpha_j \bm{v}(\xi_j)^T = \bm{0}^T$, hence $\alpha_j = 0$.
    Thus all $\alpha_j = 0$, proving that $\bm{C}_{\bm{\xi}}$ has full row rank $k-1$.
    
    Now suppose $\bm{K}(\bm{\xi}) [\bm{u}^T, \bm{v}^T]^T = \bm{0}$ for some $\bm{u} \in \mathbb{R}^{k(d+1)}$ and $\bm{v} \in \mathbb{R}^{k-1}$.
    The first block equation gives $\bm{H}(\bm{\xi}) \bm{u} + \bm{C}_{\bm{\xi}}^T \bm{v} = \bm{0}$, so $\bm{u} = -\bm{H}(\bm{\xi})^{-1} \bm{C}_{\bm{\xi}}^T \bm{v}$.
    Substituting into the second block equation $\bm{C}_{\bm{\xi}} \bm{u} = \bm{0}$ yields $\bm{C}_{\bm{\xi}} \bm{H}(\bm{\xi})^{-1} \bm{C}_{\bm{\xi}}^T \bm{v} = \bm{0}$.
    The Schur complement $\bm{S}(\bm{\xi}) := \bm{C}_{\bm{\xi}} \bm{H}(\bm{\xi})^{-1} \bm{C}_{\bm{\xi}}^T$ is symmetric positive definite: for any nonzero $\bm{v}$, the vector $\bm{C}_{\bm{\xi}}^T \bm{v}$ is nonzero (since $\bm{C}_{\bm{\xi}}$ has full row rank), and thus $\bm{v}^T \bm{S}(\bm{\xi}) \bm{v} = (\bm{C}_{\bm{\xi}}^T \bm{v})^T \bm{H}(\bm{\xi})^{-1} (\bm{C}_{\bm{\xi}}^T \bm{v}) > 0$.
    Therefore $\bm{v} = \bm{0}$, which implies $\bm{u} = \bm{0}$.
    This proves that $\bm{K}(\bm{\xi})$ is nonsingular.
\end{proof}

We now establish the convergence properties of Algorithm~\ref{alg:greedy_regression}.
Since the candidate set $\mathcal{X} = \{(x_i + x_{i+1})/2 : i = 1, \ldots, n-1\}$ is finite, the set
\begin{equation*}
\mathcal{A}\subseteq \mathcal{X}^{k-1}
\end{equation*}
of admissible interior-breakpoint configurations (those satisfying Assumption~\ref{assump:my_assumptions}) is finite.
To keep admissibility invariant under updates, we adopt the following feasible-candidate convention in Algorithm~\ref{alg:update_break_point}: if a left/right candidate would violate Assumption~\ref{assump:my_assumptions} (\textbf{A2}) or (\textbf{A4}), that candidate is treated as infeasible and assigned local objective value $+\infty$ (equivalently, excluded from the local minimization).
Hence every generated iterate remains in $\mathcal{A}$.
With a stopping rule that terminates either at a fixed point (no breakpoint updates) or upon revisiting a previously generated configuration (cycle detection), Algorithm~\ref{alg:greedy_regression} must terminate after finitely many iterations.

\begin{theorem}\label{thm:convergence}
Let $\{\bm{\xi}^{(t)}\}_{t\ge 0}$ be the sequence of breakpoint vectors generated by Algorithm~\ref{alg:greedy_regression} under the feasible-candidate convention above.
For each admissible $\bm{\xi}$, let $\bm{\theta}^\star(\bm{\xi})$ denote the unique solution of the constrained least-squares problem \eqref{eq:main2} (guaranteed by Theorem~\ref{thm:well_posed}) and define the global objective value
\begin{equation*}
F(\bm{\xi}) := \operatorname{MSE}\!\left(\bm{X}(\bm{\xi})\bm{\theta}^\star(\bm{\xi}),\,\bm{y}\right).
\end{equation*}
Let $\hat{\bm{\xi}}^{(t)}$ denote the best configuration encountered up to iteration $t$, i.e.,
\begin{equation*}
\hat{\bm{\xi}}^{(t)} \in \operatorname*{argmin}_{0\le s\le t} F(\bm{\xi}^{(s)}).
\end{equation*}
Assume the while-loop in Algorithm~\ref{alg:greedy_regression} terminates as soon as either
\begin{itemize}
\item[(a)] $\bm{\xi}^{(t+1)}=\bm{\xi}^{(t)}$ (a fixed point: no breakpoint updates), or
\item[(b)] $\bm{\xi}^{(t+1)}=\bm{\xi}^{(s)}$ for some $s\in\{0,1,\ldots,t\}$ (cycle detection).
\end{itemize}
Then:
\begin{itemize}
\item[(i)] The sequence $\{F(\hat{\bm{\xi}}^{(t)})\}_{t\ge 0}$ is monotonically non-increasing.
\item[(ii)] Algorithm~\ref{alg:greedy_regression} terminates after finitely many iterations; in particular, it performs at most $|\mathcal{A}|$ while-loop iterations (equivalently, it generates at most $|\mathcal{A}|+1$ configurations including the terminal repeated one).
\item[(iii)] If the algorithm terminates by (a) at some iteration $T$, then for each breakpoint index $j\in\{1,\ldots,k-1\}$, let
\begin{equation*}
r_-^{(T,j)},\; r^{(T,j)},\; r_+^{(T,j)}
\end{equation*}
denote the three \emph{local} MSE values computed by Algorithm~\ref{alg:update_break_point} at iteration $T$ for breakpoint $j$ (i.e., the objective values of the local two-interval surrogate problems \eqref{opti1}--\eqref{opti3}, restricted to feasible candidates).
Then one necessarily has the stationarity alternative
\begin{equation*}
r^{(T,j)} \le \min\{r_-^{(T,j)},\,r_+^{(T,j)}\}
\quad\text{or}\quad
r_-^{(T,j)} = r_+^{(T,j)} < r^{(T,j)}.
\end{equation*}
In particular, under the mild non-degeneracy condition that the tie case
$r_-^{(T,j)} = r_+^{(T,j)} < r^{(T,j)}$ never occurs, the fixed point implies
$r^{(T,j)}\le r_-^{(T,j)}$ and $r^{(T,j)}\le r_+^{(T,j)}$ for all $j$, i.e., each breakpoint is coordinate-wise locally optimal \emph{for the local two-interval surrogate objective}.
\end{itemize}
\end{theorem}

\begin{proof}
(i) By definition,
\begin{equation*}
F(\hat{\bm{\xi}}^{(t)}) = \min_{0\le s\le t} F(\bm{\xi}^{(s)}).
\end{equation*}
Since $\{0,1,\ldots,t\}\subseteq \{0,1,\ldots,t+1\}$, we have
\begin{align*}
F(\hat{\bm{\xi}}^{(t+1)})
&= \min_{0\le s\le t+1} F(\bm{\xi}^{(s)}) \\
&\le \min_{0\le s\le t} F(\bm{\xi}^{(s)})
= F(\hat{\bm{\xi}}^{(t)}).
\end{align*}
Hence $\{F(\hat{\bm{\xi}}^{(t)})\}_{t\ge 0}$ is monotonically non-increasing.

(ii) Let $\mathcal{A}$ be the set of admissible breakpoint configurations. Since $\mathcal{X}$ is finite with $|\mathcal{X}|=n-1$ and admissible configurations belong to $\mathcal{X}^{k-1}$, the set $\mathcal{A}$ is finite and satisfies $|\mathcal{A}|\le (n-1)^{k-1}$.
By the feasible-candidate convention, every iterate $\bm{\xi}^{(t)}$ lies in $\mathcal{A}$.
Assume for contradiction that the while-loop executes at least $|\mathcal{A}|+1$ iterations.
Then the first $|\mathcal{A}|$ iterations (with indices $t=0,\ldots,|\mathcal{A}|-1$) all end without triggering (a) or (b), otherwise the loop would have already stopped.
Hence, at each such iteration, $\bm{\xi}^{(t+1)}\notin\{\bm{\xi}^{(0)},\ldots,\bm{\xi}^{(t)}\}$, so the configurations
\begin{equation*}
\bm{\xi}^{(0)},\bm{\xi}^{(1)},\ldots,\bm{\xi}^{(|\mathcal{A}|)}
\end{equation*}
are pairwise distinct.
This gives $|\mathcal{A}|+1$ distinct elements in $\mathcal{A}$, impossible because $|\mathcal{A}|$ is the cardinality of $\mathcal{A}$.
Therefore the while-loop performs at most $|\mathcal{A}|$ iterations.

(iii) Assume termination occurs by (a) at iteration $T$, i.e., $\bm{\xi}^{(T+1)}=\bm{\xi}^{(T)}$.
Fix an index $j$.
At iteration $T$, Algorithm~\ref{alg:update_break_point} computes three local objective values
$r_-^{(T,j)}, r^{(T,j)}, r_+^{(T,j)}$ (restricted to feasible candidates) and then updates $\xi_j^{(T+1)}$ according to the strict-comparison rule in Algorithm~\ref{alg:update_break_point}.
Since $\xi_j^{(T+1)}=\xi_j^{(T)}$, neither the left update condition
\begin{equation*}
r_-^{(T,j)} < r^{(T,j)} \ \text{and}\ r_-^{(T,j)} < r_+^{(T,j)}
\end{equation*}
nor the right update condition
\begin{equation*}
r_+^{(T,j)} < r_-^{(T,j)} \ \text{and}\ r_+^{(T,j)} < r^{(T,j)}
\end{equation*}
can hold.
If $r^{(T,j)}\le \min\{r_-^{(T,j)},r_+^{(T,j)}\}$, we are done.
Otherwise, we must have $\min\{r_-^{(T,j)},r_+^{(T,j)}\} < r^{(T,j)}$.
Assume w.l.o.g. that $r_-^{(T,j)} < r^{(T,j)}$.
Since the left-update condition fails, we cannot have $r_-^{(T,j)} < r_+^{(T,j)}$, hence $r_-^{(T,j)}\ge r_+^{(T,j)}$.
Therefore $r_+^{(T,j)} \le r_-^{(T,j)} < r^{(T,j)}$, so in particular $r_+^{(T,j)} < r^{(T,j)}$.
If $r_+^{(T,j)} < r_-^{(T,j)}$ were strict, then the right-update condition would hold, contradicting $\xi_j^{(T+1)}=\xi_j^{(T)}$.
Hence $r_+^{(T,j)} = r_-^{(T,j)} < r^{(T,j)}$.
This proves the alternative:
$r^{(T,j)} \le \min\{r_-^{(T,j)},r_+^{(T,j)}\}$ or $r_-^{(T,j)} = r_+^{(T,j)} < r^{(T,j)}$.
\end{proof}

\begin{remark}
    One nonterminal outer iteration of Algorithm~\ref{alg:greedy_regression} consists of (i) $k-1$ calls to Algorithm~\ref{alg:update_break_point} and (ii) one global evaluation via Algorithm~\ref{alg:piecewise_regression}.
    For each breakpoint update, Algorithm~\ref{alg:update_break_point} solves three local KKT systems of size $(2d+3)\times(2d+3)$.
    Aggregating over all $k-1$ breakpoints, building the corresponding normal equations over the local subproblems costs $\mathcal{O}(n(d+1)^2)$ per iteration, and the local linear solves cost $\mathcal{O}(k(d+1)^3)$.
    The global evaluation step solves the KKT system in \eqref{eq:matrixform1} whose dimension scales with $k(d+1)+(k-1)$; solving it as a dense system would cost $\mathcal{O}((k(d+1))^3)$, while exploiting the block-diagonal structure of $\bm{X}^T\bm{X}$ and the sparsity of the continuity constraints yields a cost on the order of $\mathcal{O}(k(d+1)^3)$.
    Consequently, under the typical regime $k\ll n$ and fixed $d$, the per-iteration cost is dominated by $\mathcal{O}(n(d+1)^2)$, which becomes $\mathcal{O}(n)$ for $d=1$.
\end{remark}

\section{Experimental Results}

In this section, numerical experiments with some synthetic and real data are presented to show the effectiveness of our proposed algorithm. 
Every numerical experiment uses \MATLAB and the tests were executed on a high-performance computing system with the following specifications: an AMD Ryzen 3970X CPU running at 3.69 GHz.

\subsection{Synthetic data}

In this section, we compare the proposed method with several regression methods with synthetic data $\{(x_i, y_i)\}_{i=1}^n\subset \mathbb{R}^2$. 
The benchmark methods are as follows:

\smallskip

\noindent
\textbf{Polynomial Regression (PR)} \\
In a polynomial regression model~\citep{stigler1971optimal}, a polynomial of degree $d$ is fit to given data.
The goal is to minimize the difference between the predicted value and the actual value for each data point.
\begin{equation}
\operatorname*{minimize}_{\bm{\beta}} \sum_{i=1}^{n} \left(y_i - \sum_{j=0}^d \beta_j x_i^j\right)^2
\end{equation}

\smallskip

\noindent
\textbf{Spline Regression (SR)} \\
A spline model is a combination of polynomials that smoothly connects data points.
The optimization problem minimizes the sum of the squared errors at the data points while simultaneously minimizing the integral of the square of the second derivative to maintain the smoothness of the spline.
Here, $\lambda$ controls the trade-off between smoothness and data fit.
\begin{equation}
\operatorname*{minimize}_{f} p \sum_{i=1}^n w_i\left|y_i-f\left(x_i\right)\right|^2+(1-p) \int \lambda(t)\left|D^2 f(t)\right|^2 d t
\end{equation}
\smallskip

\noindent
\textbf{Support Vector Regression (SVR)} \\
Support vector regression~\citep{smola2004tutorial} is a regression method that aims to ensure as many data points as possible are within an $\epsilon$-tube.
Let $\phi$ be a kernel function that maps the input data to a high-dimensional feature space.
The $\psi_i$ and $\hat{\psi}_i$ ($i=1,\ldots n$) are slack variables that allow for errors for data points outside the $\epsilon$-tube.
\begin{equation}
\begin{aligned}
\operatorname*{minimize}_{w, b, \psi, \hat{\psi}} & \frac{1}{2} |w|^2 + \lambda  \sum_{i=1}^n (\psi_i + \hat{\psi}_i) \\
\text{subject to} \quad & y_i - (w \phi(x_i) + b) \leq \epsilon + \psi_i, \\
& (w \phi(x_i) + b) - y_i \leq \epsilon + \hat{\psi}_i, \\
& \psi_i, \hat{\psi}_i \geq 0,
\end{aligned}
\end{equation}
where $\lambda$ is a trade-off parameter trade-off between the flatness of model and the amount up to which deviations larger than $\epsilon$ are tolerated.\\
\smallskip

\noindent
\textbf{Decision Tree Regression (DT)} \\
In decision tree regression~\citep{loh2011classification}, the data space is partitioned into multiple bins $R_m$, and the mean value $\bar{y}_{R_m}$ within each bin is used as the predicted value.
The optimization problem is to minimize the sum of squared errors within these bins:
\begin{equation*}
\operatorname*{minimize}_{T} \sum_{m=1}^M \sum_{i \in R_m} (y_i - \bar{y}_{R_m})^2.
\end{equation*}
\smallskip

\noindent
\textbf{Gradient Boosting Regression (GB)} \\
In gradient boosting regression~\citep{friedman2001greedy}, the model $ F_0 $ is initially set to the mean of the target variable $ y $, expressed as $ F_0(x) = \bar{y} $. At each iteration $ m $, the residuals $ r_i^{(m)} $ are calculated based on the difference between the actual target values $ y_i $ and the predictions from the current ensemble model $ F_{m-1} $. This is given by $ r_i^{(m)} = y_i - F_{m-1}(x_i) $.
A weak learner $ h_m $ is then trained to predict these residuals by minimizing the sum of squared errors for the residuals:
\begin{equation*}
    h_m = \arg \min_{h} \sum_{i=1}^N (r_i^{(m)} - h(x_i))^2.
\end{equation*}
 After $ M $ iterations, the final model is expressed as:
\begin{equation*}
    F_M(x) = F_0(x) + \sum_{m=1}^M \nu \cdot h_m(x).
\end{equation*}
Here, $ \nu $ is a parameter that controls the contribution of each weak learner, helping to prevent overfitting by ensuring that the updates are gradual.
\smallskip

\noindent
\textbf{Random Forest Regression (RF)} \\
In random forest regression~\citep{breiman2001random}, multiple decision trees are trained on different subsets of the data, and their predictions are averaged to produce a final output. This method reduces overfitting and improves the model's generalization by leveraging the diversity among the trees.
\begin{equation*}
    \operatorname*{minimize}_{f} \sum_{i=1}^{n} \left( y_i - \hat{y}_i \right)^2, \quad \text{where} \quad \hat{y}_i = \frac{1}{T} \sum_{t=1}^{T} h_t(x_i)
\end{equation*}
Here, $f$ is random forest regression prediction function, given data $\{(x_i, y_i)\}_{i=1}^n\subset \mathbb{R}^2$, $\hat{y}_i$ is the predicted value for data point $i$, $T$ is the number of learning cycles, and $h_t(x_i)$ is the prediction of the $t$-th decision tree learner for $x_i$.
\\
\smallskip

\noindent
\textbf{$\bm{\ell_1}$ Trend Filter} \\
$\ell_1$ trend filter~\citep{kim2009ell_1} detects trends in data while avoiding sharp changes in trend. It does this by penalizing the sum of the absolute values of the trend changes while minimizing the absolute value of the error.
\begin{equation}
\operatorname*{minimize}_{\hat{\bm{y}}} \frac{1}{2}\|\bm{y}-\hat{\bm{y}}\|_2^2+\gamma\|\bm{D} \hat{\bm{y}}\|_1,
\end{equation}
where $\bm{y} = [y_1,\ldots,y_n]^T$ is given data, $\hat{\bm{y}} = [\hat{y}_1,\ldots,\hat{y}_n]^T$ is predicted value and $\bm{D}$ is the second-order difference matrix, i.e,
\begin{equation*}
    D=\begin{bmatrix}
    1 & -2 & 1 & & & & \\
    & 1 & -2 & 1 & & & \\
    & & \ddots & \ddots & \ddots & & \\
    & & & 1 & -2 & 1 & \\
    & & & & 1 & -2 & 1  
\end{bmatrix}\in\mathbb{R}^{(n-2)\times n}.
\end{equation*}
\smallskip

\noindent
\textbf{Adaptive Piecewise Linear Regression (APLR)} \\
Adaptive piecewise linear regression~\citep{jeong2023trend} fits linear models to segments of the data. The optimization problem is to minimize the sum of squared errors within each segment while ensuring continuity at the segment boundaries.
Let
\begin{equation*}
    F(\bm{\xi}) = \sum_{j=1}^k \sum_{x_i \in I_j}\left(p_j^d\left(x_i\right)-y_i\right)^2.
\end{equation*}

\begin{equation}
\begin{aligned}
& \operatorname{minimize} F(\bm{\xi}) \\
& \text{subject to } p_j^d\left(\xi_j\right)=p_{j+1}^d\left(\xi_j\right), j=1, \ldots, k-1
\end{aligned}
\end{equation}
Then the breakpoints $\bm{\xi}$ is updated as follows:
\begin{equation*}
\bm{\xi} = \bm{\xi} - \mu \nabla F\left( \bm{\xi}\right).
\end{equation*}
Here, $\mu$ is the gradient descent step.\\
\smallskip

\noindent
\textbf{Pruned Exact Linear Time (PELT)} \\
The pruned exact linear time algorithm~\citep{killick2012optimal} detects changes in data points, divides them into regions, and applies a linear model to each region.
The optimization problem controls the complexity of the model by assigning a penalty $\beta$ to the number of change points $k$ while minimizing the overall cost function.
\begin{equation}
\operatorname*{minimize}_{{t_1, t_2, \dots, t_k}} \left( \sum_{j=1}^{k+1} \sum_{i=t_{j-1}}^{t_j-1} (y_i - \hat{y}_i)^2 + k \beta \right)
\end{equation}

In particular, the `$\mathtt{ischange}$' function, a built-in function of \MATLAB to detect trend change points, is based on this algorithm.

To generate synthetic data $\{(x_i, y_i)\}_{1\leq i \leq 400}$ randomly, there are two processes: (i) generate a piecewise linear function $f$ with $\bm{\xi} = [1, 70, 150, 230, 300, 350, 400]^T$ and randomly generated values
$f(\xi_i) \sim \mathcal{U}\{-15,15\}$.
Here, $\mathcal{U}$ presents a discrete uniform distribution.
The value of $y_i$ at fixed $x_i=i$ is randomly generated by $y_i = f(x_i) + \epsilon_i$ with noise $\epsilon_i \sim \mathcal{N}(0,2)$.

\begin{figure}[htbp]
    \centering
    \includegraphics[width=0.85\textwidth]{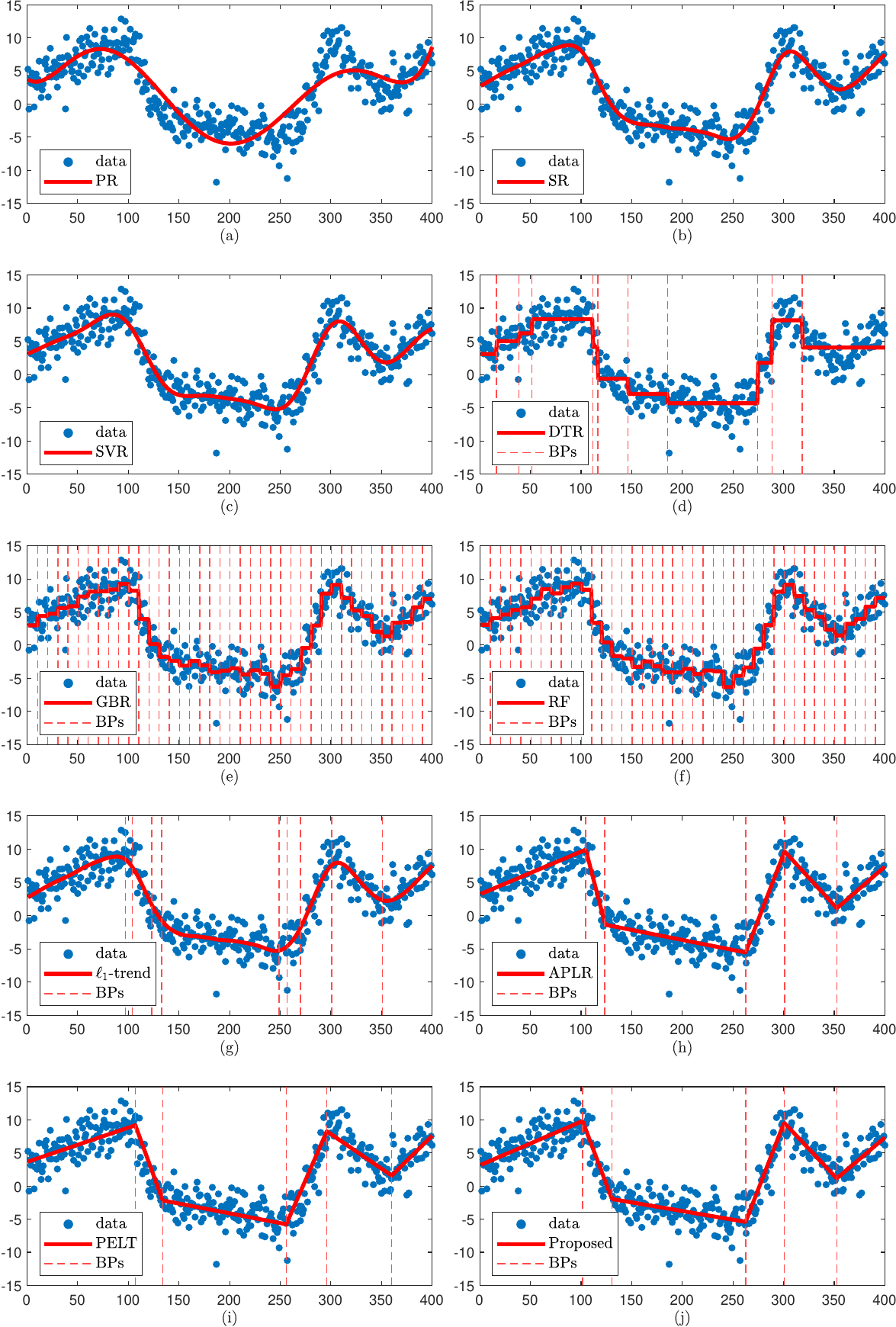}
    \caption{Results of each regression model.}
    \label{fig:compare_methods}
\end{figure}

The following evaluation metrics are introduced to compare the performance of the algorithm: mean square error~(MSE), mean absolute error~(MAE), relative absolute error~(RAE), coefficient of determination~($R^2$), and number of breakpoints~(BPs). These metrics were used to compare the performance of different algorithms.
\begin{align*}
    \operatorname{MSE}(\bm{y},\hat{\bm{y}}) &= \frac{1}{n} \sum_{i=1}^{n}(y_{i} - \hat{y}_{i})^2\\
    \operatorname{RMSE}(\bm{y},\hat{\bm{y}}) &= \sqrt{\operatorname{MSE}(\bm{y},\hat{\bm{y}})}\\
    \operatorname{MAE}(\bm{y},\hat{\bm{y}}) &= \frac{1}{n} \sum\limits_{i=1}^{n}|y_i - \hat{y}_i|\\
    \operatorname{RAE}(\bm{y},\hat{\bm{y}}) &= \frac{\sum\limits_{i=1}^{n}|y_{i} - \hat{y}_i|}{\sum\limits_{i=1}^{n}|y_{i} - \bar{\bm{y}}|}\\
    R^2 &= 1 -  \frac{\sum_{i=1}^{n}(y_i-\hat y_i)^2}{\sum_{i=1}^{n}(y_i-\bar{\bm{y}})^2}\\
    \operatorname{BPs} &= \textrm{number of breakpoints}
\end{align*}
Here, $\bar{\bm{y}}$ presents the average of $\bm{y}=[y_1,\ldots, y_n]^T$, i.e., $\bar{\bm{y}}=\frac{1}{n}\sum_{i=1}^{n}y_i$.

Note that the number of breakpoints is a significant factor in determining the balance between model complexity and the ability to capture the underlying trends. 
A higher number of breakpoints can indicate overfitting, where the model is too closely tailored to the observed data and may become unnecessarily complex. 
In contrast, too few breakpoints might result in a model that cannot adequately capture the trend within the data.

\begin{table}[htbp]
\centering
\begin{tabular}{cccccc}
\toprule
\textbf{Model} & \textbf{MAE} & \textbf{RAE} & \textbf{MSE} & \textbf{$\bm{R^2}$} & \textbf{BPs} \\
\midrule
PR & 2.1924 & 0.4837 & 7.5856 & 0.7200 & 0 \\
SR & 1.5716 & 0.3468 & 4.0950 & 0.8489 & 0 \\
SVR & 1.5857 & 0.3499 & 4.2108 & 0.8446 & 0 \\
DT & 1.6739 & 0.3693 & 4.5007 & 0.8339 & 10 \\
GB & 1.5586 & 0.3439 & 3.9687 & 0.8535 & 39 \\
RF & 1.5621 & 0.3447 & 4.0066 & 0.8521 & 39 \\
$\ell_1$ trend & 1.6146 & 0.3562 & 4.3264 & 0.8403 & 8 \\
APLR & $\bm{1.5368}$ & $\bm{0.3391}$ & 3.9545 & 0.8541 & $\bm{5}$ \\
PELT & 1.6282 & 0.3592 & 4.4442 & 0.8360 & $\bm{5}$ \\
Proposed & 1.5387 & 0.3395 & $\bm{3.9428}$ & $\bm{0.8545}$ & $\bm{5}$ \\
\bottomrule
\end{tabular}
\caption{Comparison of different regression models (Metrics from Figure~\ref{fig:compare_methods}).}
\label{tab:comparison}
\end{table}

Table~\ref{tab:comparison} compares various piecewise regression methods based on several performance metrics: MAE, RAE, MSE, $R^2$, and BPs. 
In evaluating these metrics, a few critical points emerge. 
The result of the proposed method in Table~\ref{tab:comparison} shows remarkable performance in several metrics. 
It achieves the highest $R^2$ value of 0.8545, indicating a strong fit to the data. 
In addition, it shows the lowest MSE and RAE, which are crucial indicators of fit quality on this dataset.
One of the most notable aspects of the proposed method is its balance in the number of breakpoints. The proposed method has only 5 breakpoints, it avoids the pitfall of overfitting methods like DT and RF exhibit, which use 10 and 39 breakpoints, respectively. At the same time, it overcomes the limitation of methods such as polynomial and spline, which use no breakpoints and may not capture the underlying trend accurately~(see Figure~\ref{fig:compare_methods}).

This balance highlights the strength of the proposed method in providing a flexible approach to piecewise regression. It uses enough breakpoints to capture structure while remaining parsimonious, yielding a favorable in-sample trade-off between fit quality and model complexity.

\subsection{Robustness analysis}

To further validate the robustness and consistency of the proposed algorithm, repeated experiments were conducted under varying conditions. 
Among the methods compared in Table~\ref{tab:comparison}, we select APLR and PELT as representative piecewise linear regression methods for this analysis.
APLR~\citep{jeong2023trend} is a gradient descent-based approach that iteratively updates breakpoint locations, while PELT~\citep{killick2012optimal} is a dynamic programming-based method widely used for change point detection.
These two methods represent fundamentally different algorithmic strategies and serve as meaningful baselines for evaluating the proposed method.

The synthetic data were generated using the same piecewise linear function with breakpoints $\bm{\xi} = [1, 100, 130, 260, 300, 350, 400]^T$ and values $[3, 10, -2, -5, 9, 2, 6]^T$.
Each experiment was repeated $50$ times with different random noise realizations, and the results are reported as mean $\pm$ standard error.

Table~\ref{tab:robustness_n} presents the comparison results under different sample sizes with fixed noise level $\sigma = 2$.
The proposed method consistently outperforms both APLR and PELT across all sample sizes, achieving approximately $15\%$ and $11\%$ improvement in MSE, respectively.
As the sample size increases, all methods show more stable performance, as indicated by decreasing standard errors.

\begin{table}[htbp]
\centering
\caption{Performance comparison under different sample sizes ($\sigma=2$, $k=5$, 50 repetitions).}
\label{tab:robustness_n}
\begin{tabular}{llcc}
\toprule
$n$ & Method & MSE & $R^2$ \\
\midrule
400 & \textbf{Proposed} & $\bm{3.96 \pm 0.04}$ & $\bm{0.846 \pm 0.002}$ \\
    & APLR & $4.68 \pm 0.05$ & $0.818 \pm 0.002$ \\
    & PELT & $4.46 \pm 0.06$ & $0.827 \pm 0.002$ \\
\midrule
800 & \textbf{Proposed} & $\bm{3.90 \pm 0.03}$ & $\bm{0.848 \pm 0.001}$ \\
    & APLR & $4.57 \pm 0.03$ & $0.822 \pm 0.001$ \\
    & PELT & $4.17 \pm 0.03$ & $0.837 \pm 0.001$ \\
\midrule
1600 & \textbf{Proposed} & $\bm{3.97 \pm 0.02}$ & $\bm{0.846 \pm 0.001}$ \\
     & APLR & $4.66 \pm 0.02$ & $0.819 \pm 0.001$ \\
     & PELT & $4.13 \pm 0.02$ & $0.840 \pm 0.001$ \\
\bottomrule
\end{tabular}
\end{table}

Table~\ref{tab:robustness_sigma} shows the performance comparison under different noise levels with fixed sample size $n = 400$.
As expected, the performance of all methods degrades with increasing noise level.
However, the proposed method maintains its advantage across all noise conditions, demonstrating its robustness to noise.
At low noise level ($\sigma = 1$), the proposed method achieves an $R^2$ value of $0.957$, while at high noise level ($\sigma = 4$), the $R^2$ value drops to $0.581$.
The proposed method consistently achieves the lowest MSE and highest $R^2$ compared to both APLR and PELT across all noise levels.

\begin{table}[htbp]
\centering
\caption{Performance comparison under different noise levels ($n=400$, $k=5$, 50 repetitions).}
\label{tab:robustness_sigma}
\begin{tabular}{llcc}
\toprule
$\sigma$ & Method & MSE & $R^2$ \\
\midrule
1 & \textbf{Proposed} & $\bm{0.97 \pm 0.01}$ & $\bm{0.957 \pm 0.001}$ \\
  & APLR & $1.64 \pm 0.02$ & $0.927 \pm 0.001$ \\
  & PELT & $1.11 \pm 0.02$ & $0.951 \pm 0.001$ \\
\midrule
2 & \textbf{Proposed} & $\bm{3.91 \pm 0.04}$ & $\bm{0.848 \pm 0.001}$ \\
  & APLR & $4.63 \pm 0.04$ & $0.821 \pm 0.001$ \\
  & PELT & $4.42 \pm 0.06$ & $0.829 \pm 0.002$ \\
\midrule
4 & \textbf{Proposed} & $\bm{15.86 \pm 0.16}$ & $\bm{0.581 \pm 0.004}$ \\
  & APLR & $16.26 \pm 0.15$ & $0.571 \pm 0.004$ \\
  & PELT & $18.18 \pm 0.29$ & $0.520 \pm 0.007$ \\
\bottomrule
\end{tabular}
\end{table}

\subsection{Real data}
A linear regression may not be a proper model for real data.
Rather, piecewise linear regression can provide the flexibility to capture abrupt changes or trend changes at specific points in the data.
In this section, the proposed method is applied to a variety of real-world data and compared with the results of other piecewise regression methods.
The real financial data set and the real epidemic data set are considered. 

\subsubsection*{S\&P 500 index data}

\begin{figure*}[ht]
    \centering
    \includegraphics[width=0.9\textwidth]{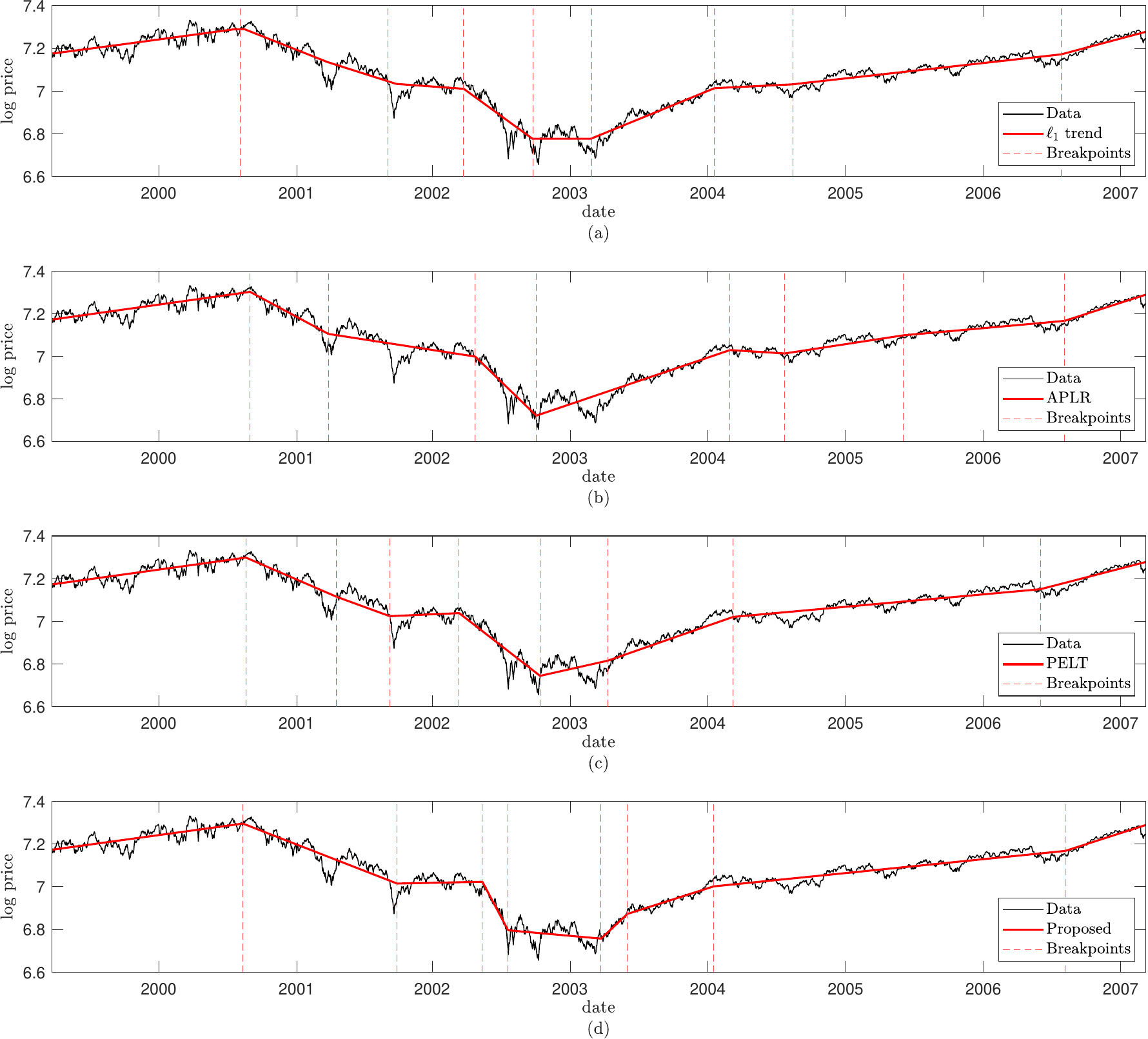}
    \caption{Comparison of $\ell_1$ trend filter, APLR, PELT and proposed method for S\&P 500 data.}
    \label{fig:SnP500}
\end{figure*}

S\&P 500 index data are the most representative and widely traded indices in their respective markets, namely the New York Stock Exchange.
The dataset is generated every business day from March 25, 1999, to March 9, 2007, and regression is performed on the logarithm of the adjusted closing price.
The logarithm transformation is widely used in financial analysis to stabilize variance and make data more suitable for certain statistical modeling.

$\ell_1 $trend filter, APLR, and PELT, which are widely known techniques in the field of piecewise regression and inflection point detection, were selected as comparison subjects.
Set the parameter $\gamma=100$ for the $\ell_1$ trend filter.
The number of breakpoints for APLR and PELT is set to $10$ using the heuristic approach, and the number of initial breakpoints for Algorithm \ref{alg:optimal_BP_algorithm} was set to $15$.

\begin{table}[htbp]
\centering
\begin{tabular}{cccccc}
\toprule
\textbf{Model} & \textbf{MAE} & \textbf{RAE} & \textbf{RMSE} & \textbf{$\bm{R^2}$} & \textbf{BPs} \\
\midrule
$\ell_1$ trend & 0.0239 & 0.2018 & 0.0314 & 0.9548 & 8 \\
APLR & 0.0242 & 0.2041 & 0.0327 & 0.9509 & 8 \\
PELT & 0.0257 & 0.2169 & 0.0332 & 0.9496 & 8 \\
\textbf{Proposed} & $\bm{0.0228}$ & $\bm{0.1921}$ & $\bm{0.0299}$ & $\bm{0.9592}$ & $\bm{8}$ \\
\bottomrule
\end{tabular}
\caption{Performance comparison of $\ell_1$ trend filter, APLR, PELT, proposed method for S\&P 500 data (see Figure~\ref{fig:SnP500}).}
\label{tab:SnP500}
\end{table}

Table~\ref{tab:SnP500} presents a performance comparison of four methods: $\ell_1$ trend filter, APLR, PELT, and a proposed method evaluated on S\&P 500 data across MAE, RAE, RMSE, $R^2$, and the number of breakpoints.
All methods identify $8$ breakpoints, ensuring comparability. 
The proposed method outperforms the others, achieving the lowest MAE ($0.0228$), RAE ($0.1921$), and RMSE ($0.0299$), along with the highest $R^2$ value ($0.9592$), indicating a better fit to the data. 
This shows the effectiveness of the proposed method in capturing the underlying trend in the data (see Figure~\ref{fig:SnP500}).

\subsubsection*{COVID-19 case data}

\begin{figure*}[ht]
    \centering
    \includegraphics[width=0.9\textwidth]{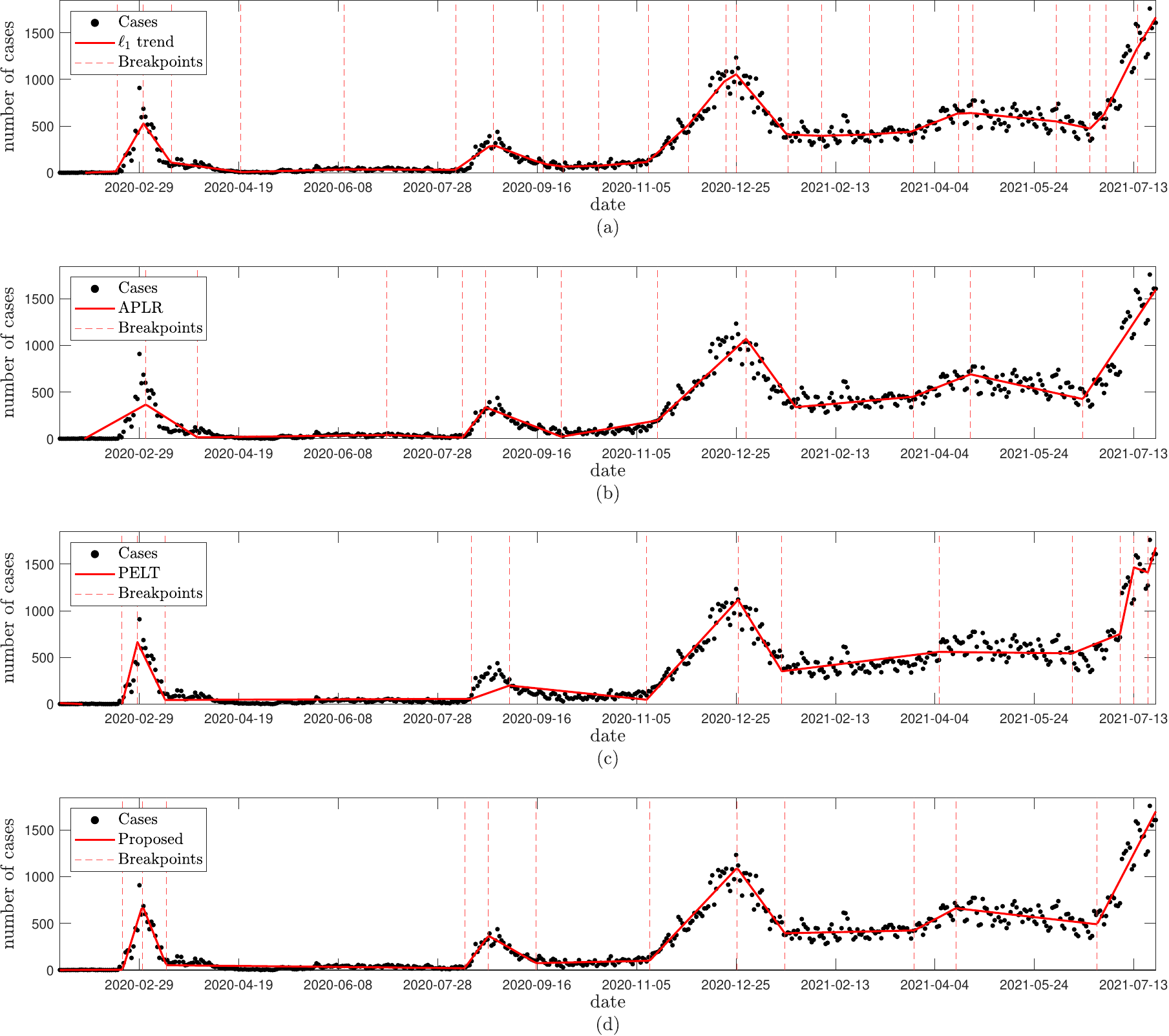}
    \caption{Comparison of $\ell_1$ trend filter, APLR, PELT and proposed method for COVID-19 cases dataset.}
    \label{fig:covid}
\end{figure*}

The second real-world dataset is the COVID-19 data provided by the Korea Disease Control and Prevention Agency (KDCA)~\citep{KDCA2} from January 20, 2020 to July 24, 2021.
This period preceded the predominance of the delta variant of the COVID-19 virus, and the South Korean government implemented a phased approach to quarantine during this time, including social distancing and mandatory masks~\citep{lee2022forecasting}.

Regression is performed on the number of confirmed cases (positive cases for real-time reverse transcription polymerase chain reaction test).
$\ell_1$ trend filter, APLR, PELT, and the proposed method for the dataset are considered.
The black dots represent daily cases, and the red line represents the regression model fitted to the data. 
The red dashed lines represent breakpoints where a change in trend is observed~(see Figure~\ref{fig:covid}).

\begin{table}[htbp]
    \centering
    \begin{tabular}{lccccc}
        \toprule
\textbf{Model} & \textbf{MAE} & \textbf{RAE} & \textbf{RMSE} & \textbf{$\bm{R^2}$} & \textbf{BPs} \\
        \midrule
        $\ell_1$ trend & \textbf{46.6322} & \textbf{0.1696} & 71.6204 & 0.9557 & 24 \\
        APLR & 60.8801 & 0.2214 & 92.4369 & 0.9262 & 12 \\
        PELT & 63.4558 & 0.2308 & 88.4348 & 0.9325 & 13 \\
        \textbf{Proposed} & 47.5149 & 0.1728 & \textbf{70.8662} & \textbf{0.9566} & \textbf{12} \\
        \bottomrule
    \end{tabular}
    \caption{Performance comparison of $\ell_1$ trend filter, APLR, PELT, proposed method for Korean COVID-19 data (see Figure~\ref{fig:covid}).}
    \label{tab:covid}
\end{table}

Table~\ref{tab:covid} compares among the models using several metrics.
Although the $\ell_1$ trend filter achieves the lowest MAE and RAE, the proposed method exhibits the best RMSE and $R^2$ values, indicating a superior overall model performance in terms of fit to the data.
The proposed method also identifies fewer breakpoints which is $12$, suggesting a more parsimonious model compared to $\ell_1$ trend filter, which identifies $24$ breakpoints.
This indicates that the proposed method effectively balances the trade-off between model complexity and fit quality. 
The ability of the model to follow the main trends without overfitting to short-term fluctuations is crucial to accurately capturing the dynamics of the pandemic. 

\section{Conclusion}

In this study, we proposed a novel piecewise regression algorithm that effectively identifies breakpoints to minimize errors. 
Through numerical experiments using synthetic and real data, we show that the proposed method has high performance in piecewise polynomial regression. Furthermore, the algorithm can determine the optimal number of breakpoints.

In future work, the process of detecting breakpoints in piecewise regression should consider using reinforcement learning that takes into account long-term rewards. 
Although current algorithms focus on short-term gains and run the risk of converging to local minima, reinforcement learning may lead to more optimized results. 
This approach is expected to increase the accuracy and stability of the models and improve their adaptability to different types of data and their robustness.

\section*{Disclosure statement}
The authors report there are no competing interests to declare.

\section*{Funding}
The work of T.~Kim, H.~Lee, and H.~Choi was supported by the National Research Foundation of Korea (NRF) grant funded by the Korea government (MSIT) (No. 2022R1A5A1033624~\&~RS-2024-00342939).
The work of T.~Kim was additionally supported by the National Research Foundation of Korea (NRF) grant funded by the Korea government (MSIT) (No. RS-2025-25436769).
M.~Kim was supported by the Basic Science Research Program and the G-LAMP program (Global-Learning \& Academic research institution for masters and PhD students and postdocs) through the National Research Foundation of Korea (NRF) funded by the Ministry of Education (No. RS-2023-00248103~\&~RS-2023-00301914).

\section*{Data availability statement}
The data that support the findings of this study are available from the corresponding author upon reasonable request.
Publicly available datasets (S\&P 500, COVID-19) were used in this study.

\bibliography{RL}

@article{kim2009ell_1,
  title={$\ell_1$ trend filtering},
  author={Kim, Seung-Jean and Koh, Kwangmoo and Boyd, Stephen and Gorinevsky, Dimitry},
  journal={SIAM review},
  volume={51},
  number={2},
  pages={339--360},
  year={2009},
  publisher={SIAM}
}

@article{jeong2023trend,
  title={Trend filtering by adaptive piecewise polynomials},
  author={Jeong, Juyoung and Jung, Yoon Mo and Kim, Soo Hyun and Yun, Sangwoon},
  journal={Communications in Nonlinear Science and Numerical Simulation},
  volume={116},
  pages={106866},
  year={2023},
  publisher={Elsevier}
}

@article{tibshirani2014adaptive,
  title={Adaptive piecewise polynomial estimation via trend filtering},
  author={Tibshirani, Ryan J},
  journal={The Annals of Statistics},
  volume={42},
  number={1},
  pages={285},
  year={2014},
  publisher={Institute of Mathematical Statistics}
}

@article{muggeo2003estimating,
  title={Estimating regression models with unknown break-points},
  author={Muggeo, Vito MR},
  journal={Statistics in medicine},
  volume={22},
  number={19},
  pages={3055--3071},
  year={2003},
  publisher={Wiley Online Library}
}

@article{bai1997estimation,
  title={Estimation of a change point in multiple regression models},
  author={Bai, Jushan},
  journal={Review of Economics and Statistics},
  volume={79},
  number={4},
  pages={551--563},
  year={1997},
  publisher={MIT Press}
}

@book{quarteroni2010numerical,
  title={Numerical mathematics},
  author={Quarteroni, Alfio and Sacco, Riccardo and Saleri, Fausto},
  volume={37},
  year={2010},
  publisher={Springer Science \& Business Media}
}

@article{fearnhead2007line,
  title={On-line inference for multiple changepoint problems},
  author={Fearnhead, Paul and Liu, Zhen},
  journal={Journal of the Royal Statistical Society Series B: Statistical Methodology},
  volume={69},
  number={4},
  pages={589--605},
  year={2007},
  publisher={Oxford University Press}
}

@article{bai1998estimating,
  title={Estimating and testing linear models with multiple structural changes},
  author={Bai, Jushan and Perron, Pierre},
  journal={Econometrica},
  pages={47--78},
  year={1998},
  publisher={JSTOR}
}

@article{wagner2002segmented,
  title={Segmented regression analysis of interrupted time series studies in medication use research},
  author={Wagner, Anita K and Soumerai, Stephen B and Zhang, Fang and Ross-Degnan, Dennis},
  journal={Journal of clinical pharmacy and therapeutics},
  volume={27},
  number={4},
  pages={299--309},
  year={2002},
  publisher={Wiley Online Library}
}

@article{guan2018polynomial,
  title={Polynomial time algorithms and extended formulations for unit commitment problems},
  author={Guan, Yongpei and Pan, Kai and Zhou, Kezhuo},
  journal={IISE transactions},
  volume={50},
  number={8},
  pages={735--751},
  year={2018},
  publisher={Taylor \& Francis}
}

@article{hwangbo2018spline,
  title={Spline model for wake effect analysis: Characteristics of a single wake and its impacts on wind turbine power generation},
  author={Hwangbo, Hoon and Johnson, Andrew L and Ding, Yu},
  journal={IISE Transactions},
  volume={50},
  number={2},
  pages={112--125},
  year={2018},
  publisher={Taylor \& Francis}
}

@book{ding2019data,
  title={Data science for wind energy},
  author={Ding, Yu},
  year={2019},
  publisher={Chapman and Hall/CRC}
}

@article{muriel2004capacitated,
  title={Capacitated multicommodity network flow problems with piecewise linear concave costs},
  author={Muriel, Ana and Munshi, Farhad N},
  journal={IIE Transactions},
  volume={36},
  number={7},
  pages={683--696},
  year={2004},
  publisher={Taylor \& Francis}
}

@article{gunnerud2010oil,
  title={Oil production optimization---A piecewise linear model, solved with two decomposition strategies},
  author={Gunnerud, Vidar and Foss, Bjarne},
  journal={Computers \& Chemical Engineering},
  volume={34},
  number={11},
  pages={1803--1812},
  year={2010},
  publisher={Elsevier}
}

@article{tomal2021bayesian,
  title={A Bayesian piecewise linear model for the detection of breakpoints in housing prices},
  author={Tomal, Jabed H and Rahman, Hafizur},
  journal={Metron},
  volume={79},
  number={3},
  pages={361--381},
  year={2021},
  publisher={Springer}
}

@article{greene2015improved,
  title={Improved statistical analysis of pre-and post-treatment patient-reported outcome measures (PROMs): the applicability of piecewise linear regression splines},
  author={Greene, ME and Rolfson, O and Garellick, G and Gordon, M and Nemes, S},
  journal={Quality of Life Research},
  volume={24},
  pages={567--573},
  year={2015},
  publisher={Springer}
}

@article{vieth1989fitting,
  title={Fitting piecewise linear regression functions to biological responses},
  author={Vieth, Elisabeth},
  journal={Journal of applied physiology},
  volume={67},
  number={1},
  pages={390--396},
  year={1989}
}

@article{wu2022synchronous,
  title={A synchronous multiple change-point detecting method for manufacturing process},
  author={Wu, Zhenyu and Li, Yanting and Hu, Lanye},
  journal={Computers \& Industrial Engineering},
  volume={169},
  pages={108114},
  year={2022},
  publisher={Elsevier}
}

@article{bai2003computation,
  title={Computation and analysis of multiple structural change models},
  author={Bai, Jushan and Perron, Pierre},
  journal={Journal of applied econometrics},
  volume={18},
  number={1},
  pages={1--22},
  year={2003},
  publisher={Wiley Online Library}
}

@article{chiou2019identifying,
  title={Identifying multiple changes for a functional data sequence with application to freeway traffic segmentation},
  author={Chiou, Jeng-Min and Chen, Yu-Ting and Hsing, Tailen},
  journal={The Annals of Applied Statistics},
  volume={13},
  number={3},
  pages={1430--1463},
  year={2019},
  publisher={JSTOR}
}

@article{fryzlewicz2014wild,
 author = {Piotr Fryzlewicz},
 journal = {The Annals of Statistics},
 number = {6},
 pages = {2243--2281},
 publisher = {Institute of Mathematical Statistics},
 title = {WILD BINARY SEGMENTATION FOR MULTIPLE CHANGE-POINT DETECTION},
 urldate = {2024-06-17},
 volume = {42},
 year = {2014}
}

@article{killick2012optimal,
  title={Optimal detection of changepoints with a linear computational cost},
  author={Killick, Rebecca and Fearnhead, Paul and Eckley, Idris A},
  journal={Journal of the American Statistical Association},
  volume={107},
  number={500},
  pages={1590--1598},
  year={2012},
  publisher={Taylor \& Francis}
}

@article{truong2020selective,
  title={Selective review of offline change point detection methods},
  author={Truong, Charles and Oudre, Laurent and Vayatis, Nicolas},
  journal={Signal Processing},
  volume={167},
  pages={107299},
  year={2020},
  publisher={Elsevier}
}

@misc{KDCA2,
  author = {{Korea Disease Control and Prevention Agency (KDCA)}},
  title = {KDCA Open Data Portal},
  howpublished = "\url{https://dportal.kdca.go.kr/pot/index.do}",
  note = " [accessed 18 June 2024]",
  year={2021},
}

@article{lee2022forecasting,
  title={Forecasting COVID-19 cases by assessing control-intervention effects in Republic of Korea: a statistical modeling approach},
  author={Lee, Hyojung and Jang, Geunsoo and Cho, Giphil},
  journal={Alexandria Engineering Journal},
  volume={61},
  number={11},
  pages={9203--9217},
  year={2022},
  publisher={Elsevier}
}

@article{dell2022generalizations,
  title={Generalizations of the constrained mock-Chebyshev least squares in two variables: Tensor product vs total degree polynomial interpolation},
  author={Dell'Accio, Francesco and Di Tommaso, Filomena and Nudo, Federico},
  journal={Applied Mathematics Letters},
  volume={125},
  pages={107732},
  year={2022},
  publisher={Elsevier}
}

@book{boyd2018introduction,
  title={Introduction to applied linear algebra: vectors, matrices, and least squares},
  author={Boyd, Stephen and Vandenberghe, Lieven},
  year={2018},
  publisher={Cambridge university press}
}

@article{dell2024polynomial,
  title={Polynomial approximation of derivatives through a regression--interpolation method},
  author={Dell'Accio, Francesco and Nudo, Federico},
  journal={Applied Mathematics Letters},
  volume={152},
  pages={109010},
  year={2024},
  publisher={Elsevier}
}

@article{stigler1971optimal,
  title={Optimal experimental design for polynomial regression},
  author={Stigler, Stephen M},
  journal={Journal of the American Statistical Association},
  volume={66},
  number={334},
  pages={311--318},
  year={1971},
  publisher={Taylor \& Francis}
}

@article{smola2004tutorial,
  title={A tutorial on support vector regression},
  author={Smola, Alex J and Sch{\"o}lkopf, Bernhard},
  journal={Statistics and computing},
  volume={14},
  pages={199--222},
  year={2004},
  publisher={Springer}
}

@article{friedman2001greedy,
  title={Greedy function approximation: a gradient boosting machine},
  author={Friedman, Jerome H},
  journal={Annals of statistics},
  pages={1189--1232},
  year={2001},
  publisher={JSTOR}
}

@article{loh2011classification,
  title={Classification and regression trees},
  author={Loh, Wei-Yin},
  journal={Wiley interdisciplinary reviews: data mining and knowledge discovery},
  volume={1},
  number={1},
  pages={14--23},
  year={2011},
  publisher={Wiley Online Library}
}

@article{breiman2001random,
  title={Random forests},
  author={Breiman, Leo},
  journal={Machine learning},
  volume={45},
  pages={5--32},
  year={2001},
  publisher={Springer}
}

@article{tunc2021column,
  title={A column generation based heuristic algorithm for piecewise linear regression},
  author={Tunc, Huseyin and Gen{\c{c}}, Burkay},
  journal={Expert Systems with Applications},
  volume={171},
  pages={114539},
  year={2021},
  publisher={Elsevier}
}

@article{yang2016mathematical,
  title={Mathematical programming for piecewise linear regression analysis},
  author={Yang, Lingjian and Liu, Songsong and Tsoka, Sophia and Papageorgiou, Lazaros G},
  journal={Expert systems with applications},
  volume={44},
  pages={156--167},
  year={2016},
  publisher={Elsevier}
}

\end{document}